\documentclass[11pt]{article}

\pdfoutput=1

 \usepackage{natbib}
\usepackage{hyperref}
\usepackage{graphicx}
\graphicspath{{../figures/}}

\usepackage{subcaption}
\usepackage{float}

\usepackage[colorinlistoftodos,prependcaption,textsize=tiny]{todonotes}

\usepackage{amsmath}
 \usepackage{amssymb}
\usepackage{mhequ}

\newcommand{\pr}{\text{Pr}}
\newcommand{\rulesep}{\vrule height -1ex\ }

\newcommand{\be}{\begin{equation*}
  \begin{aligned}}

\newcommand{\ee}{ \end{aligned}
\end{equation*}
}

\newcommand{\bel}{\begin{equation}
  \begin{aligned}}

\newcommand{\eel}{ \end{aligned}
\end{equation}
}

\newtheorem{theorem}{Theorem}

\newtheorem{proof}{Proof}

\begin{document}

\title{{Latent Simplex Position Model:\\
High Dimensional Multi-view Clustering \\with  Uncertainty Quantification}}
 \def\spacingset#1{\renewcommand{\baselinestretch}%
 {#1}\small\normalsize} \spacingset{1}
\author{ Leo L. Duan\thanks{Department of Statistics, University of Florida, Gainesville, FL, email: li.duan@ufl.edu}
     }
%
\maketitle

{\bf Abstract:}
High dimensional data often contain multiple facets, and several clustering patterns can co-exist under different variable subspaces, also known as the views. While multi-view clustering algorithms were proposed, the uncertainty quantification remains difficult --- a particular challenge is in the high complexity of estimating the cluster assignment probability under each view, and sharing information among views. In this article, we propose an approximate Bayes approach --- treating the similarity matrices generated over the views as rough first-stage estimates for the co-assignment probabilities; in its Kullback-Leibler neighborhood, we obtain a refined low-rank matrix, formed by the pairwise product of simplex coordinates. Interestingly, each simplex coordinate directly encodes the cluster assignment uncertainty. For multi-view clustering, we let each view draw a  parameterization from a few candidates, leading to dimension reduction. With high model flexibility, the estimation can be efficiently carried out as a continuous optimization problem, hence enjoys gradient-based computation. The theory establishes the connection of this model to a random partition distribution under multiple views. Compared to single-view clustering approaches, substantially more interpretable results are obtained when clustering brains from a human traumatic brain injury study, using high-dimensional gene expression data.

\vskip 12pt


\vfill

\newpage


\section{Introduction}

High dimensional data are becoming increasingly common in areas such as genomics, computer vision, neuroscience, etc. They are characterized by the ambient dimension substantially larger than the sample size $n$. When clustering such data, canonical solutions tend to focus on finding one particular clustering pattern. For example, one idea is to use variable selection method to identify a small subset of variables with large discriminability, then using them as input for clustering algorithms \citep{law2003feature,tadesse2005bayesian,hoff2006model,witten2010framework}; another idea is to reduce the dimension onto latent linear subspaces, often via a Gaussian mixture model with low-rank covariance structure \citep{ghahramani1996algorithm}; recent work extends this to the variational autoencoder for nonlinear dimension reduction \citep{dilokthanakul2016deep}. These methods have been successful when there is only one clear clustering result in the data. However, as high dimension data often contain multiple facets of the observations, it is more natural to consider more than one clustering patterns --- that is, different subspaces of variables can correspond to distinct clustering results. As a result, focusing on one clustering pattern --- or, `single-view' is often inadequate.

There has been active literature motivated for `multi-view clustering'. Since there are two distinct definitions of this concept, to be clear, we will focus on the one finding multiple clustering patterns, as opposed to the other aiming for one consensus based on multiple data sources. Within our scope, early work includes combining random projection and spectral clustering to obtain clusters on a randomly projected space, repeating this several times to produce multiple clustering patterns \citep{fern2003random}; using regularization framework by running clustering algorithm in each view, while minimizing the cross-view divergence \citep{kumar2011co,joshi2016renyi}.

As clusters tend to overlap, there is often substantial uncertainty in clustering. A major interest is on the randomness of cluster assignment, characterized by a categorical distribution. In the canonical single-view setting, one typically relies on the Bayesian framework by assigning a model-based likelihood \citep{Fraley:2002bg}. This requires putting a parametric assumption on each within-cluster distribution, then estimating the posterior of cluster assignment via the Markov-chain Monte Carlo (MCMC) algorithm. Among the Bayesian multi-view clustering literature \citep{niu2010multiple,li2011bayesian, kirk2012bayesian,lock2013bayesian,niu2013iterative, mo2017fully}, \cite{guan2010variational} and \cite{niu2012nonparametric} use the Indian Buffet Process to combine relevant variables into several groups, and in each group, they use a Gaussian mixture model to carry out clustering. These Bayesian models give a generative perspective for the multi-view data; however, there are two major challenges in practice: 1) assigning a within-cluster distribution is prone to misspecifying the model, which leads to breakdown of the parameter estimation \citep{hennig2004breakdown} and uncontrolled growth in the number of clusters \citep{miller2018robust}; 2) the MCMC computation suffers from a critically slow convergence/mixing as the dimension grows, limiting its high dimension application. For the former, it was recently shown that modeling the pairwise divergence has much better robustness compared to the original data \citep{duan2018bayesian}; for the latter, in general, it has become increasingly popular to replace sampling with an optimization-based approximation for the posterior distribution \citep{el2012bayesian}.

In this article, we are motivated for an approximate Bayes approach that allows for a direct estimation of the cluster assignment and co-assignment probabilities, within the scope of having several distinct clustering patterns. This is inspired by the resemblance between a similarity matrix and a cluster graph, hence the former can be considered as a noisy version of the latter. In the community detection literature, one often learns a low-rank representation for each data point as the latent position in Euclidean/Stiefel space \citep{hoff2002latent}, then cluster the coordinates into communities \citep{handcock2007model}. Instead of going through two modeling stages, we put the latent coordinates directly on the probability simplex, describing the probabilities for cluster assignment and allowing gradient-descent optimization; in the meantime, each co-assignment matrix is a random draw out of only a few candidate parameterizations, leading to dimension reduction and information sharing among views.

\section{Method}

\begin{figure}[H]
 \begin{subfigure}[t]{0.32\textwidth}
 \centering
       \includegraphics[width=1\linewidth]{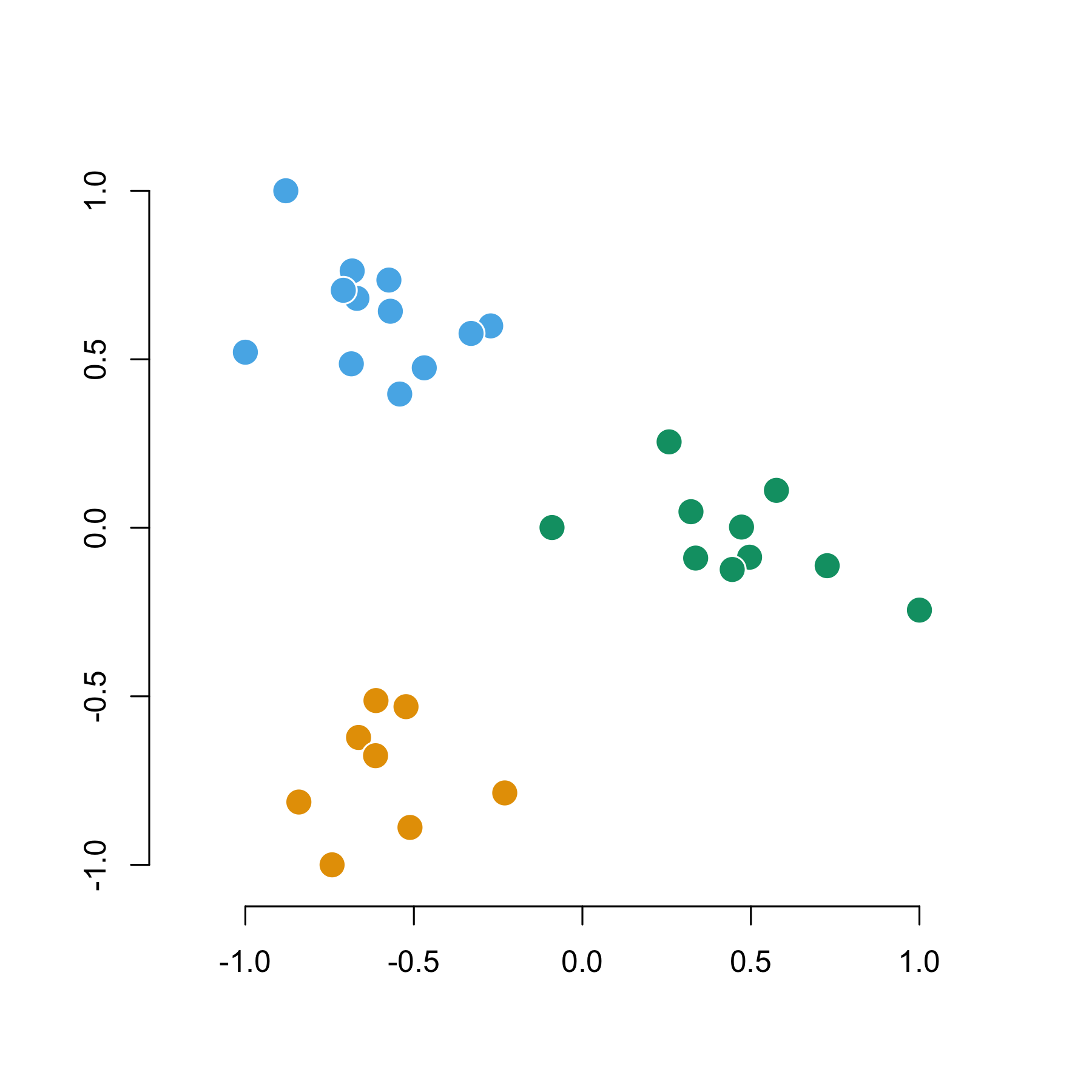}
             \caption{$y^{(1)}$}
 \end{subfigure}
 \begin{subfigure}[t]{0.32\textwidth}
 \centering
       \includegraphics[width=1\linewidth]{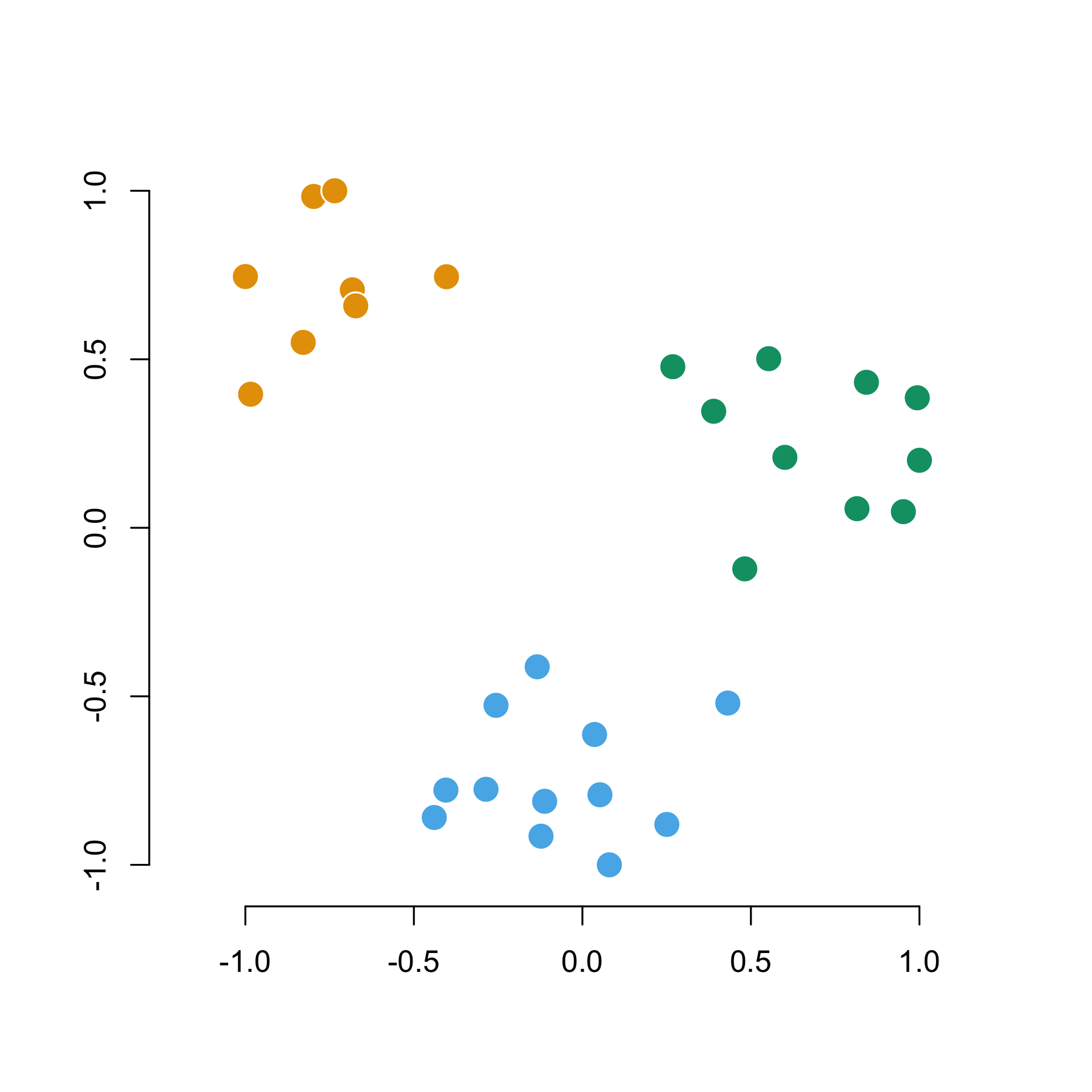}
             \caption{$y^{(2)}$}
 \end{subfigure}
\begin{subfigure}[t]{0.32\textwidth}
 \centering
       \includegraphics[width=1\linewidth]{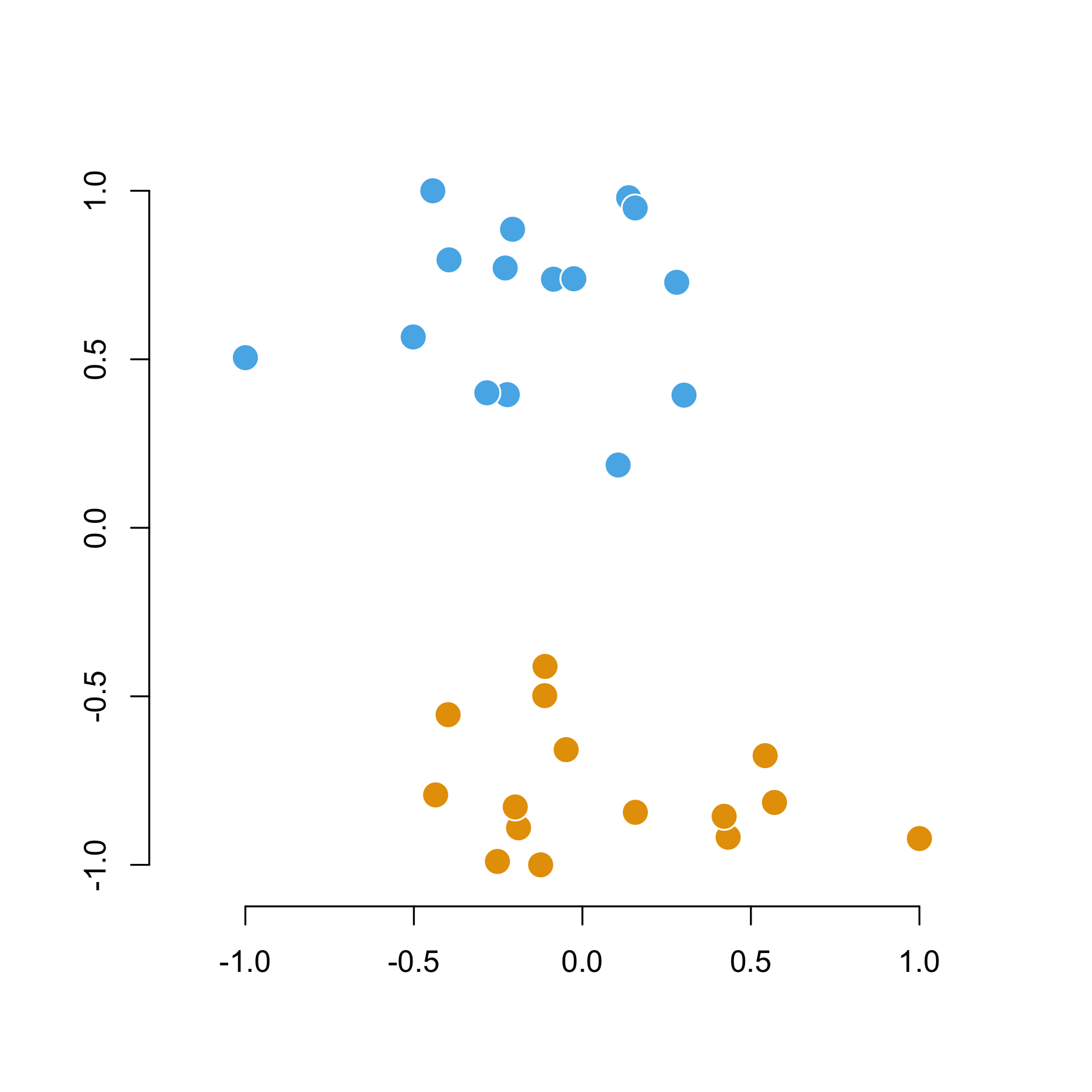}
             \caption{$y^{(3)}$}
 \end{subfigure}
 \caption{Simulated data under three views (each view is
$\mathbb{R}^2$). The colors represent the true cluster assignments in each view. \label{fig:running_example1}}
 \end{figure}

Let $y_i \in \mathcal{Y}$ be the data over $i=1,\ldots,n$. Each $y_i$ is a multi-dimensional vector, and we can separate its elements into groups of sub-vectors $(y_i^{(1)}, y_i^{(2)}, \ldots, y_i^{(V)})$. We will now call the subspace $\mathcal Y^{(v)}$ for  $y_i^{(v)}$ as a `view' of the data. To help explain the idea, we use a running example of simulated data over $3$ views (each in $\mathbb{R}^2$). Figure~\ref{fig:running_example1} shows the scatter plots.

Assuming the views are given, our goal is to obtain clustering for each view, in particular, a discrete label  $c^{(v)}_i \in\{1,\ldots,g\}$ corresponding to the cluster assignment for each $y_i^{(v)}$.  Equivalently, we can focus on the co-assignment for each pair of data
\be
z^{(v)}_{i,j} = 1 (c^{(v)}_i = c^{(v)}_j),
\ee
where $1(E)$ is the indicator function taking $1$ if $E$ is true, otherwise taking $0$.

\subsection{Latent Simplex Position Model}
\begin{figure}[H]
 \begin{subfigure}[t]{0.32\textwidth}
 \centering
       \includegraphics[width=1\linewidth]{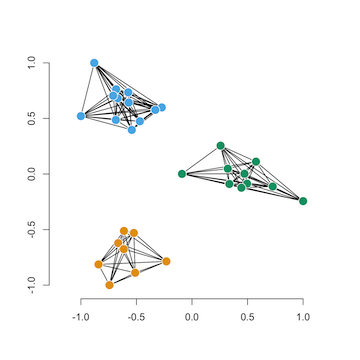}
             \caption{$G^{(1)}$}
 \end{subfigure}
 \begin{subfigure}[t]{0.32\textwidth}
 \centering
       \includegraphics[width=1\linewidth]{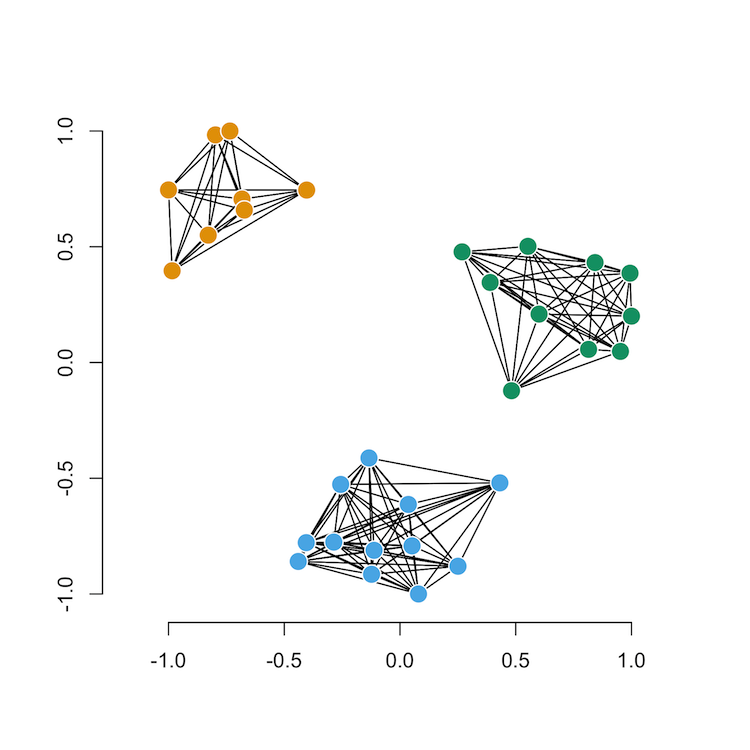}
             \caption{$G^{(2)}$}
 \end{subfigure}
\begin{subfigure}[t]{0.32\textwidth}
 \centering
       \includegraphics[width=1\linewidth]{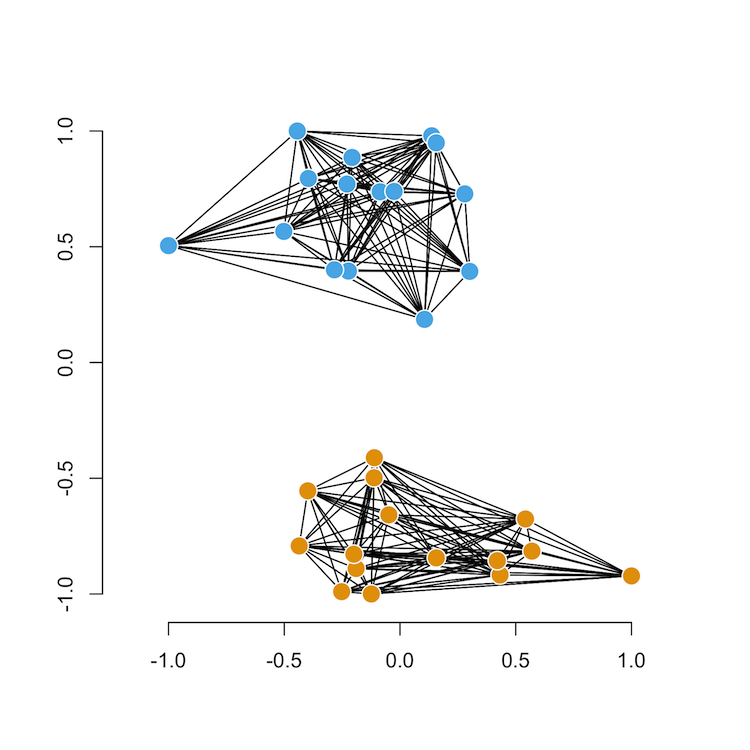}
             \caption{$G^{(3)}$}
 \end{subfigure}
 \caption{The oracle cluster graphs of the simulated data under three views:  each of $G^{(1)}$ and $G^{(2)}$ has three disconnected sub-graphs; $G^{(3)}$ has only two
sub-graphs.\label{fig:running_example2}}
 \end{figure}

Treating $Z^{(v)}=\{z_{i,j}^{(v)}\}_{i,j}$ as an adjacency matrix, we can form a cluster graph:  $G^{(v)}=(\mathcal{N},\mathcal{E}^{(v)})$, with $\mathcal N=\{1,\ldots,n\}$ and $\mathcal{E}^{(v)}= \{ e^{(v)}_{i,j} :z^{(v)}_{i,j}=1\}$. In this graph, each cluster forms a complete sub-graph (all pairs of nodes within are connected), and the sub-graphs are disconnected. Figure~\ref{fig:running_example2} plots the cluster graph for each view.

In order to handle a large number of views, we consider the following generative process: assuming there are $d$ ways to parameterize the distribution for $Z^{(v)}$, in each view, we draw one of $d$ candidate parameterizations; then we proceed to draw the cluster assignments and form a cluster graph.
\be
& x^{(v)}  \;\;\stackrel{iid}{\sim} \text{Categorical}(
\lambda_{1}, \ldots,\lambda_{d}
), \\
& c^{(v)}_{i} \mid x^{(v)}=l \;\;\; \stackrel {indep} {\sim} \text{Categorical}(
w_{i,1}^{(l)}, \ldots, w_{i,g}^{(l)}),\\
& z^{(v)}_{i,j}  = 1(c^{(v)}_{i} =c^{(v)}_{j} ),
 \ee
where $(\lambda_{1}, \ldots,\lambda_{d}) \in \Delta^{d-1}$ and $(w_{i,1}^{(l)}, \ldots, w_{i,g}^{(l)})\in \Delta^{g-1}$; $\Delta^{g-1}=\{(\nu_1,\ldots, \nu_g): \nu_k\ge 0 ,\sum_{k=1}^g\nu_k=1 \}$  is the probability simplex. Equivalently,
\bel\label{eq:cluster_graph_generative}
&         z^{(v)}_{i,j} \mid p^{(v)}_{i,j},x^{(v)}  \sim \text{Bernoulli}(p^{(v)}_{i,j}),
\\
& p^{(v)}_{i,j} = \sum_{l=1}^d 1(x^{(v)}=l) \sum_{k=1}^g w^{(l)}_{i,k}
w^{(l)}_{j,k}.
\eel

 In addition, we could further consider the data as generated from
$
 y_i^{(v)} \mid  c_{i}^{(v)}=k \sim \mathcal F_{v,k},
$
with $\mathcal F_{v,k}$ a certain distribution. However, $\mathcal F_{v,k}$ is often unknown and challenging to estimate. Instead, we will focus on a pairwise transform as a surrogate for $y^{(v)}_{i}$'s.

\begin{figure}[t]
 \begin{subfigure}[t]{0.32\textwidth}
 \centering
       \includegraphics[width=1\linewidth]{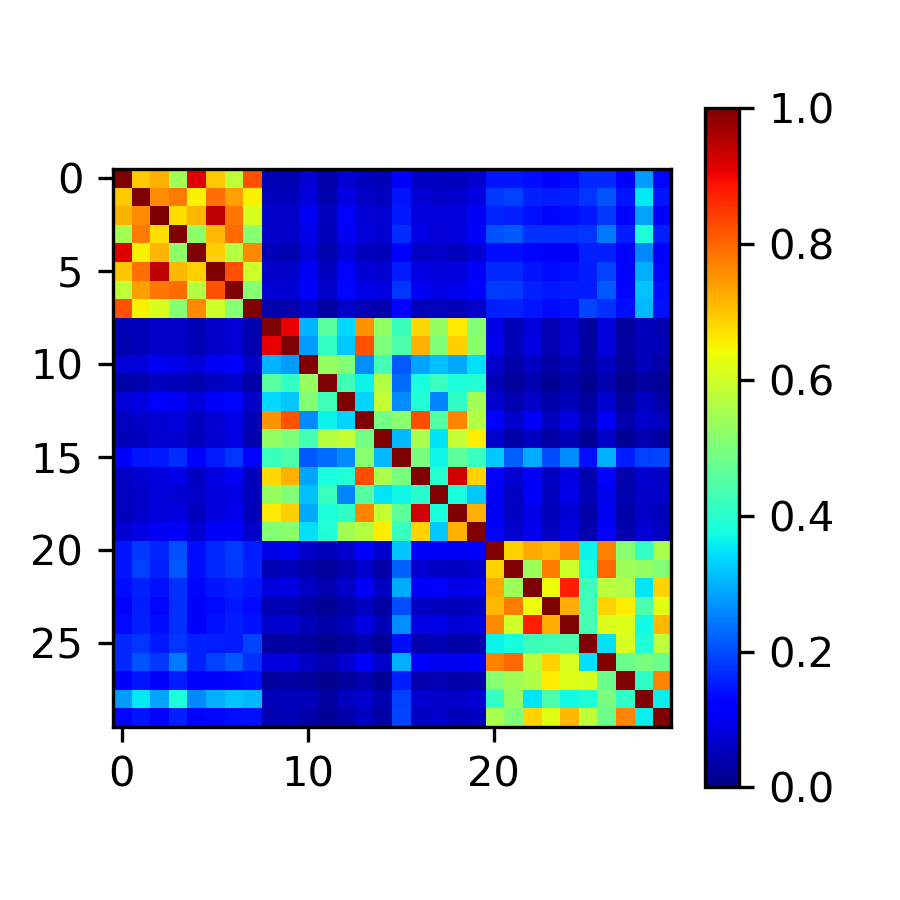}
             \caption{$S^{(1)}$}
 \end{subfigure}
 \begin{subfigure}[t]{0.32\textwidth}
 \centering
       \includegraphics[width=1\linewidth]{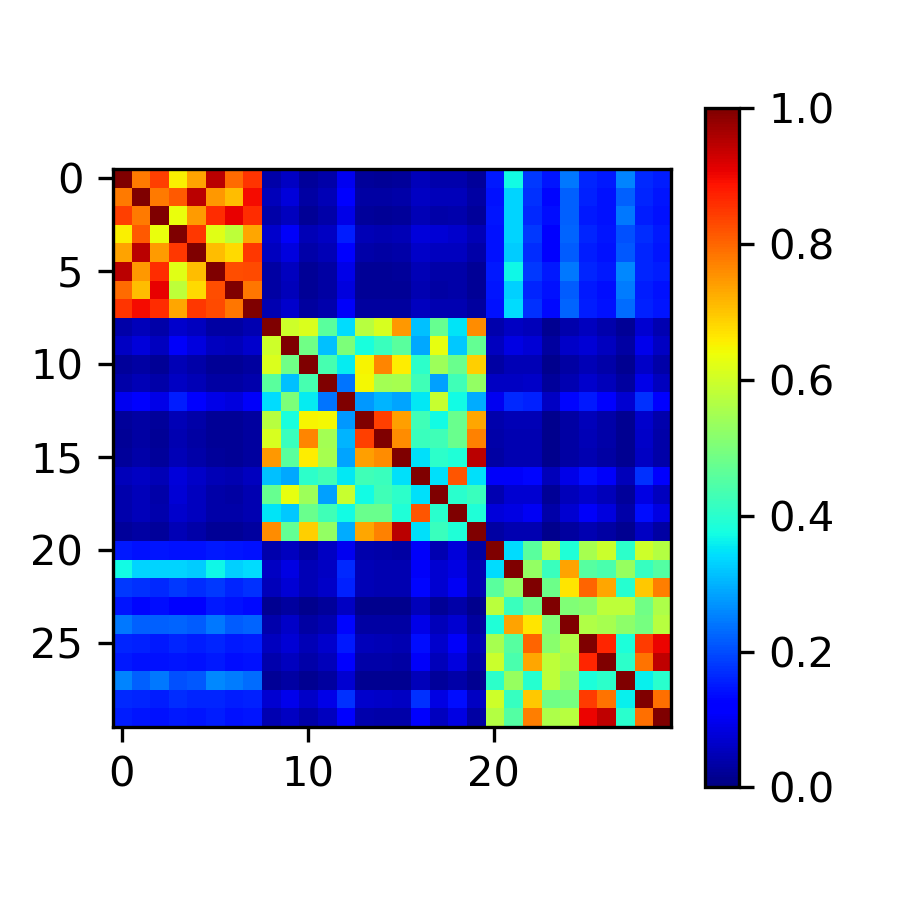}
             \caption{$S^{(2)}$}
 \end{subfigure}
\begin{subfigure}[t]{0.32\textwidth}
 \centering
       \includegraphics[width=1\linewidth]{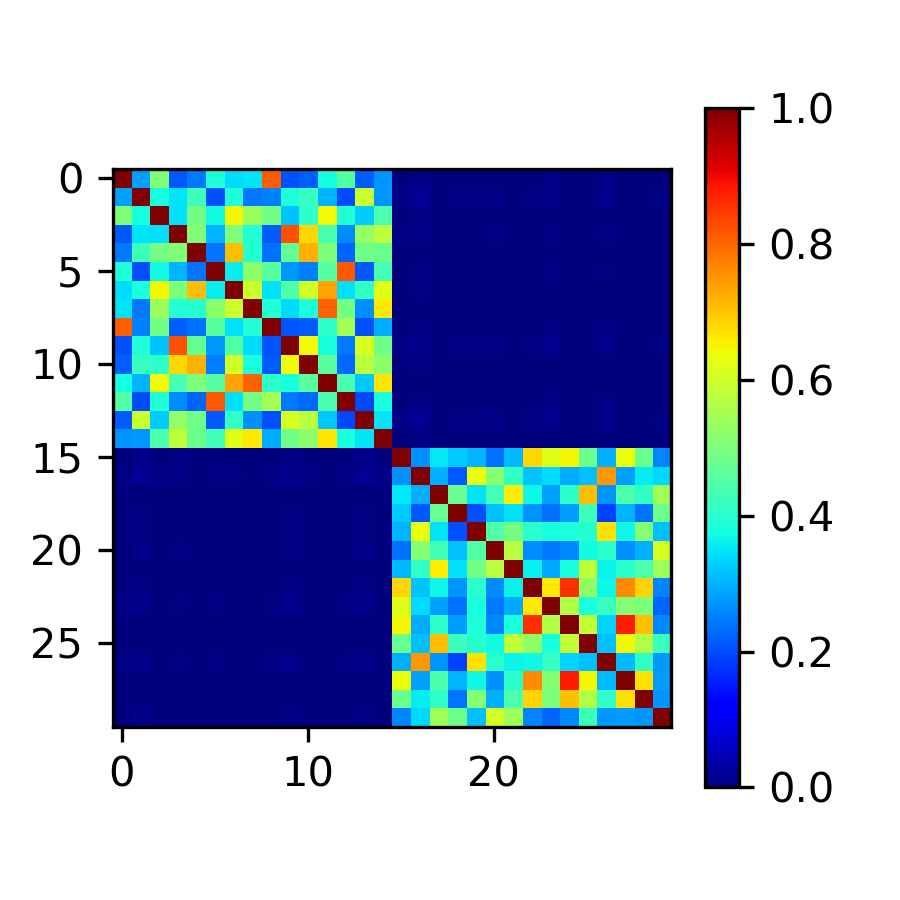}
             \caption{$S^{(3)}$}
 \end{subfigure}
  \caption{Similarity as an approximation for $\pr\big( z^{(v)}_{i,j}  =1 \mid y_i^{(v)}, y_j^{(v)} \big)$ under three views. \label{fig:running_example4}}
 \end{figure}

Intuitively, if $y_i^{(v)}$ and $y_j^{(v)}$ are close to each other, it is more likely that they are from the same $\mathcal F_{v,k}$. In machine learning, we have the similarity score to quantify such a proximity:
\bel\label{eq:empircal_est}
&  s^{(v)}_{i,j}=\mathcal{K}(y_i^{(v)},y_j^{(v)}),
\eel
where $\mathcal{K}$ is a positive semi-definite kernel that maps to $(0,1)$. This can be taken as an approximate
\bel\label{eq:approx}
s^{(v)}_{i,j}\approx\pr(z^{(v)}_{i,j}=1 ).
\eel
We will also use matrix notations $S^{(v)}=\{s^{(v)}_{i,j}\}_{i,j}$ an $n\times n$ matrix,
 $P^{(v)}=\{p^{(v)}_{i,j}\}_{i,j}$ an $n\times n$ matrix, and $W^{(l)}=\{w^{(l)}_{i,k}\}_{i,k}$ an $n\times g$ matrix.
In this article, we use a popular similarity $s^{(v)}_{i,j}=\exp(-\|
y_i^{(v)}-y_j^{(v)}\|
/b^{(v)}_{i,j})$ with $b^{(v)}_{i,j}>0$ the local bandwidth parameter  formed by the row quantiles in $S^{(v)}$, according
to \cite{zelnik2005self}. Figure~\ref{fig:running_example4} plots the similarity matrices
computed from  the simulated data.

In order to connect \eqref{eq:approx} with \eqref{eq:cluster_graph_generative}, we propose a generative model for $S^{(v)}$
\bel\label{eq:pseudolikelihood1}
& s^{(v)}_{i,j} \mid p^{(v)}_{i,j} \sim H(p^{(v)}_{i,j}),
\eel
where $H$ is a distribution such that $s^{(v)}_{i,j} $ is a noisy version of $p^{(v)}_{i,j}$. To choose its density $h$, because both $s^{(v)}_{i,j} $ and $p^{(v)}_{i,j}$ are Bernoulli probabilities, we use a pseudo-likelihood based on the Kullback-Leibler divergence (we will justify this choice in the theory section)
\bel\label{eq:pseudolikelihood2}
& h(s^{(v)}_{i,j}; p^{(v)}_{i,j}) \propto \exp \big[ -KL \big(p^{(v)}_{i,j}\|s^{(v)}_{i,j}\big) \big],\\
& KL \big(p^{(v)}_{i,j}\|s^{(v)}_{i,j}\big)
=   p^{(v)}_{i,j}
\log \frac{p^{(v)}_{i,j} }{  s^{(v)}_{i,j}}
+  (1-p^{(v)}_{i,j})   \log \frac{1-p^{(v)}_{i,j} }{  1-s^{(v)}_{i,j}
}.
\eel

 \begin{figure}[t]
 \begin{subfigure}[t]{0.32\textwidth}
 \centering
       \includegraphics[width=1\linewidth]{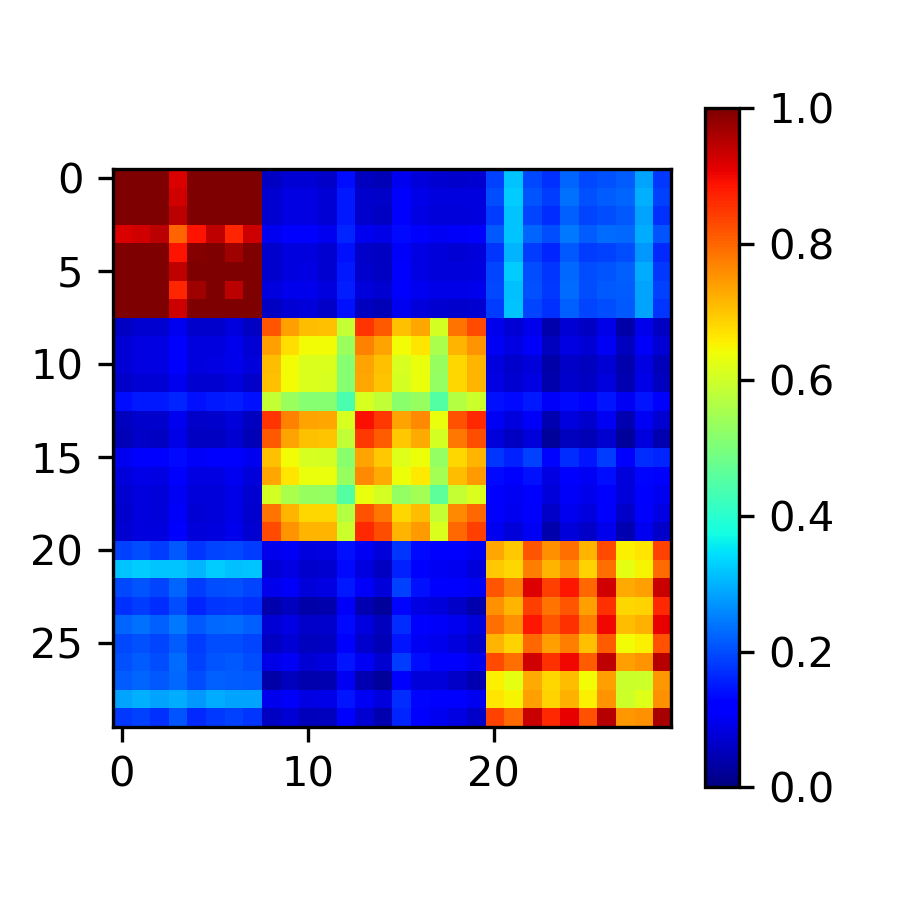}
             \caption{$P^{(1)}$}
 \end{subfigure}
 \begin{subfigure}[t]{0.32\textwidth}
 \centering
       \includegraphics[width=1\linewidth]{cluster_graphon2}
             \caption{$P^{(2)}$}
 \end{subfigure}
 \begin{subfigure}[t]{0.32\textwidth}
 \centering
       \includegraphics[width=1\linewidth]{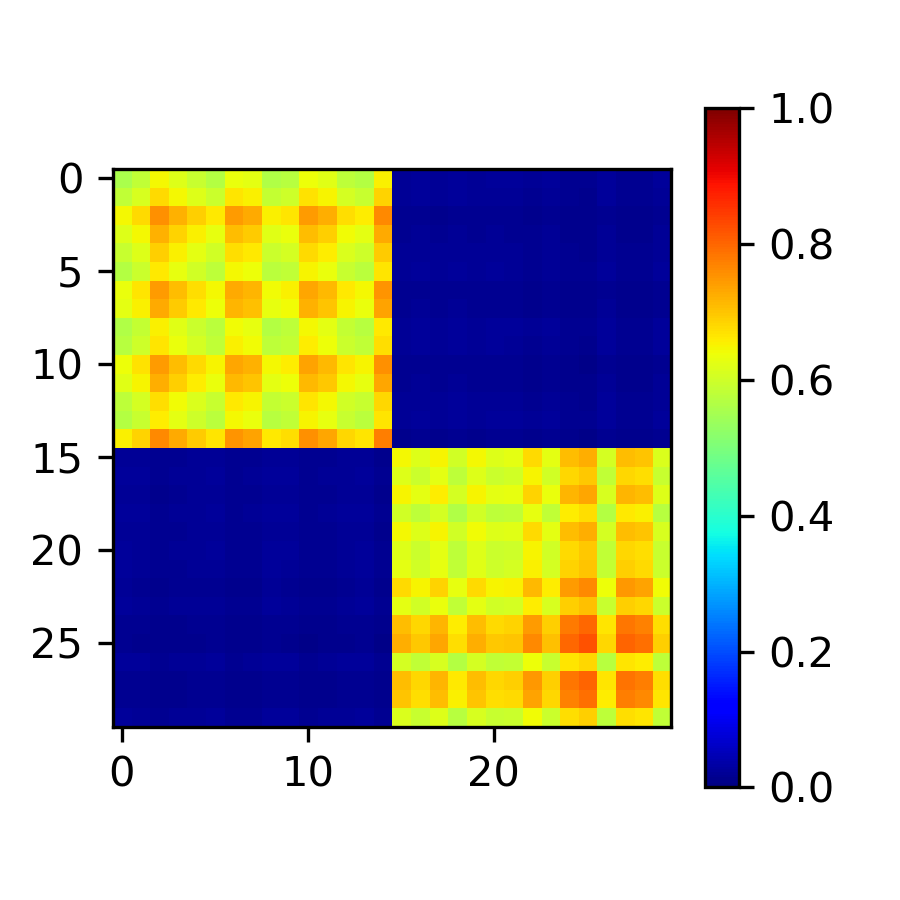}
             \caption{$P^{(3)}$}
 \end{subfigure}
   \caption{The matrix $P^{(v)}$ provides a low-rank smoothing for $S^{(v)}$: each pixel of the heatmap represents the marginal probability $\pr(z_{i,j}^{(v)}=1)$. The first two matrices have the same parameterization due to the same $x^{(1)}=x^{(2)}$. \label{fig:running_example3} }
 \end{figure}


Using \eqref{eq:pseudolikelihood1}, we can obtain the posterior distribution $\Pi( P^{(v)} \mid S^{(v)})$, as a surrogate for
$\Pi( P^{(v)} \mid y_1^{(v)} ,\ldots, y_n^{(v)})$. Although it is an approximation, a key benefit is that the posterior mode of $(P^{(v)} \mid S^{(v)})$ directly estimates the clustering uncertainty $\pr(c^{(v)}_i= c^{(v)}_j)$; and the mode of $(W^{(x^{(v)})} \mid S^{(v)})$ estimates $\pr(c^{(v)}_i=k)$.
And note that
\be
S^{(v)} \approx W^{(x^{(v)})} W^{(x^{(v)}) \rm T} = P^{(v)},
\ee
where the right hand side has the rank less or equal to $(g-1)$, providing a low-rank smoothing. Therefore, the optimization for the mode is close to a simple matrix factorization, hence it is computationally more efficient, compared to the costly MCMC algorithm.

Lastly, if we consider a `similarity graph' with $S^{(v)}$ as its adjacency matrix, then  the $i$th row of $W^{(x^{(v)})}$
\be
(w^{(x^{(v)})}_{i,1}, \ldots, w^{(x^{(v)})}_{i,g}) \in \Delta^{g-1}
\ee
is a latent position for the node $i$. Therefore, our model is
 a special case of the latent position model \citep{hoff2002latent}; and we name it as the latent simplex position (LSP) model.

\subsection{Regularization in Overfitted Model}
Often we do not know the minimally needed number of clusters $g_0$ (or the `truth'), instead, we assign an overfitted model with an over-specified $g\gg g_0$. It is useful to consider regularization: suppose that the cluster $k$ is redundant, we can use some regularization term to force the $k$th column in $W^{(l)}$ to be close to zero, $w_{i,k}^{(l)}=\pr(c_i^{(l)}=k) \approx 0$ for $i=1,\ldots,n$. Similarly, we want to over-specify $d$ and use regularization to force some redundant $\lambda_l\approx 0$.

We use a regularized loss function
\bel\label{eq:simplex_reg}
RegLoss =     &KL(P\|S)  +   n\sum_{l=1}^d R\big ( W^{(l)}\big)    + \sum_{l=1}^d T(\lambda_l),
  \eel
  with $KL(P\|S)= \sum_{v=1}^V\sum_{j<i} KL \big(p^{(v)}_{i,j}\|s^{(v)}_{i,j}\big)$ and two regularization terms $R\big ( W^{(l)}\big ) $ and $T(\lambda_l)$.

\begin{figure}[H]
\begin{subfigure}[t]{0.48\textwidth}
\begin{subfigure}[t]{0.46\textwidth}
\centering
      \includegraphics[width=1\linewidth]{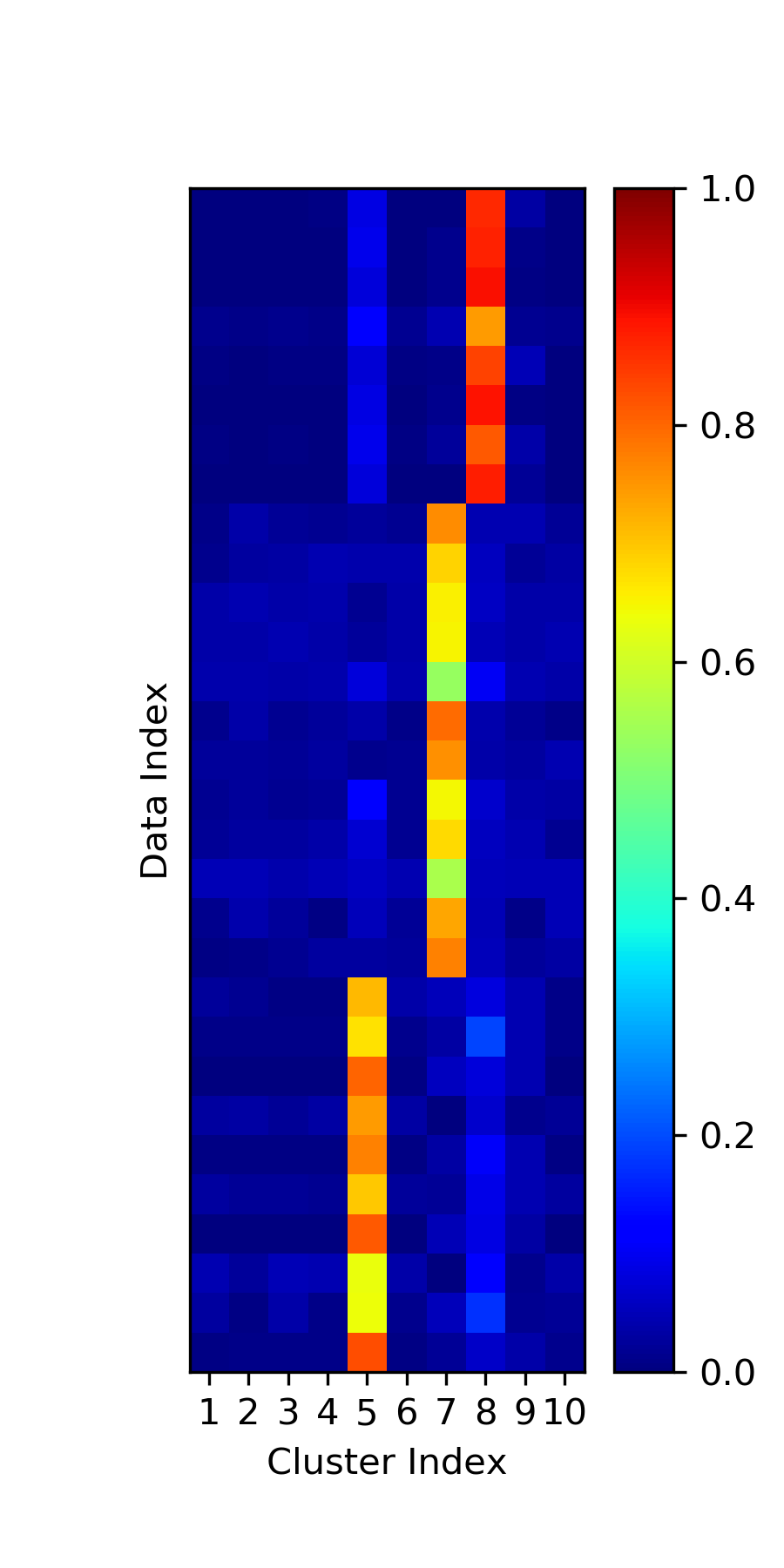}
\end{subfigure}
\begin{subfigure}[t]{0.46\textwidth}
\centering
      \includegraphics[width=1\linewidth]{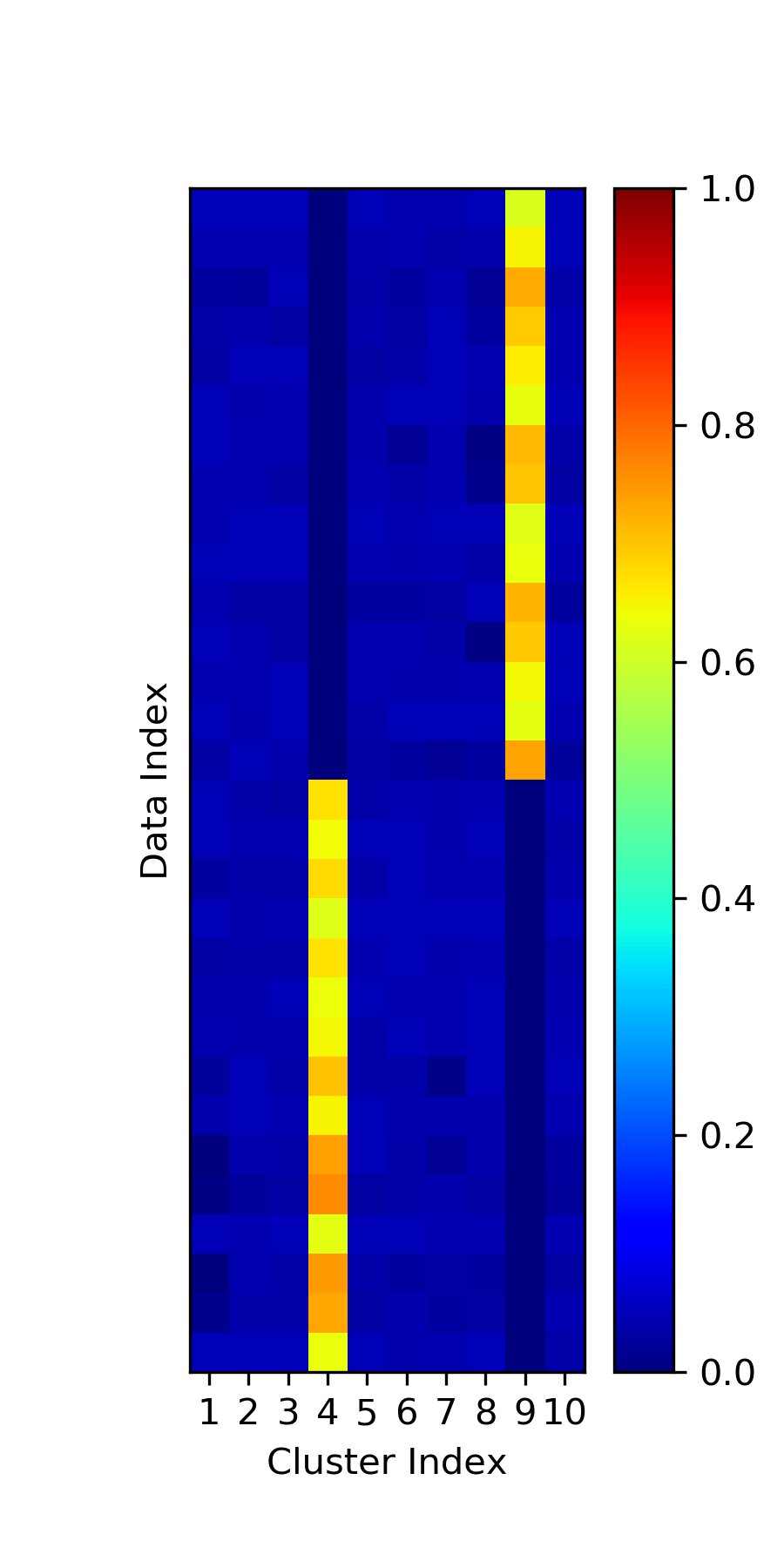}
      \end{subfigure}
      \caption{Estimated $W^{(l)}$ matrix using group regularization.}\
\end{subfigure}
\begin{subfigure}[t]{0.48\textwidth}
\begin{subfigure}[t]{0.46\textwidth}
\centering
      \includegraphics[width=1\linewidth]{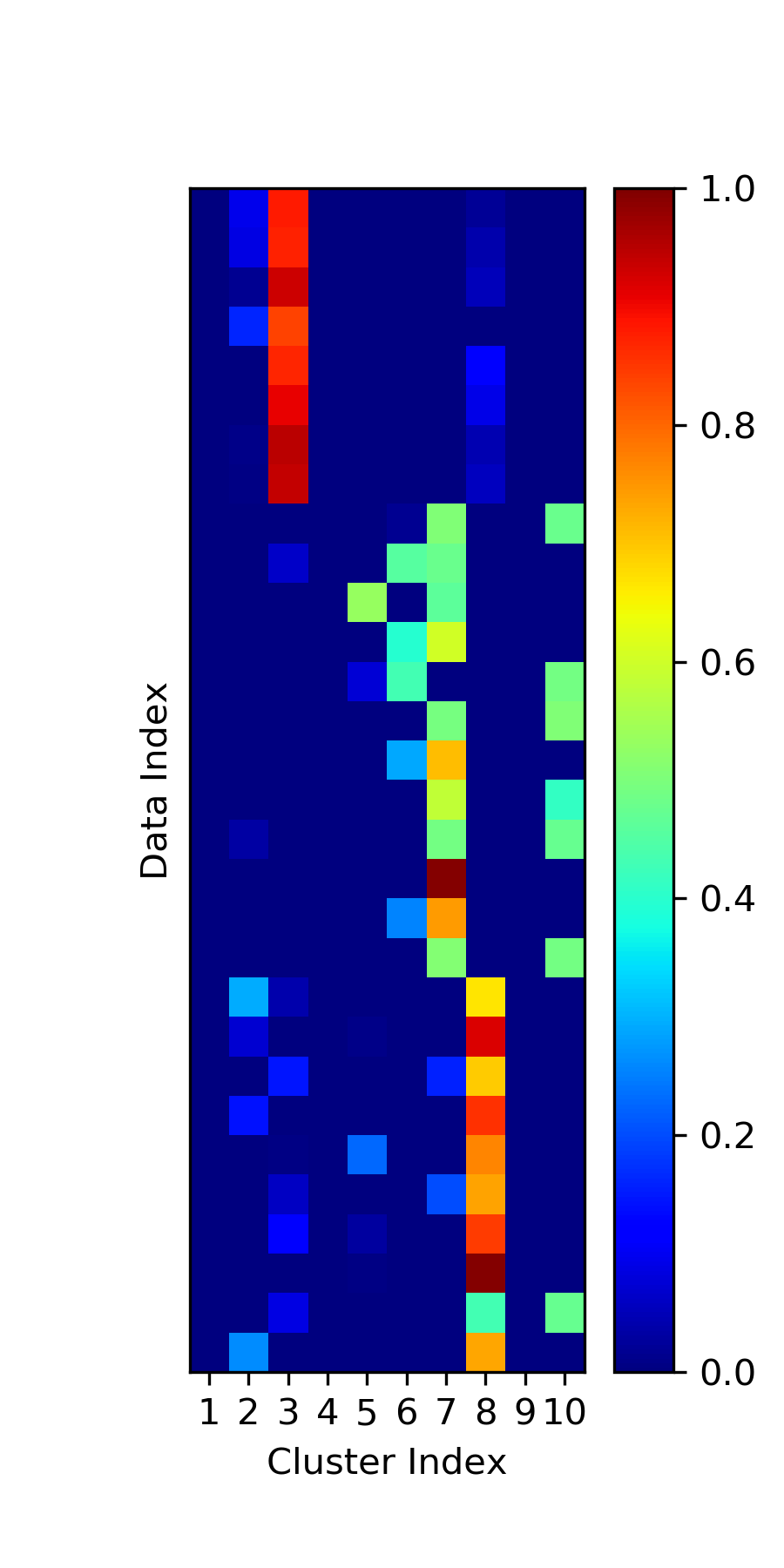}
\end{subfigure}
\begin{subfigure}[t]{0.46\textwidth}
\centering
      \includegraphics[width=1\linewidth]{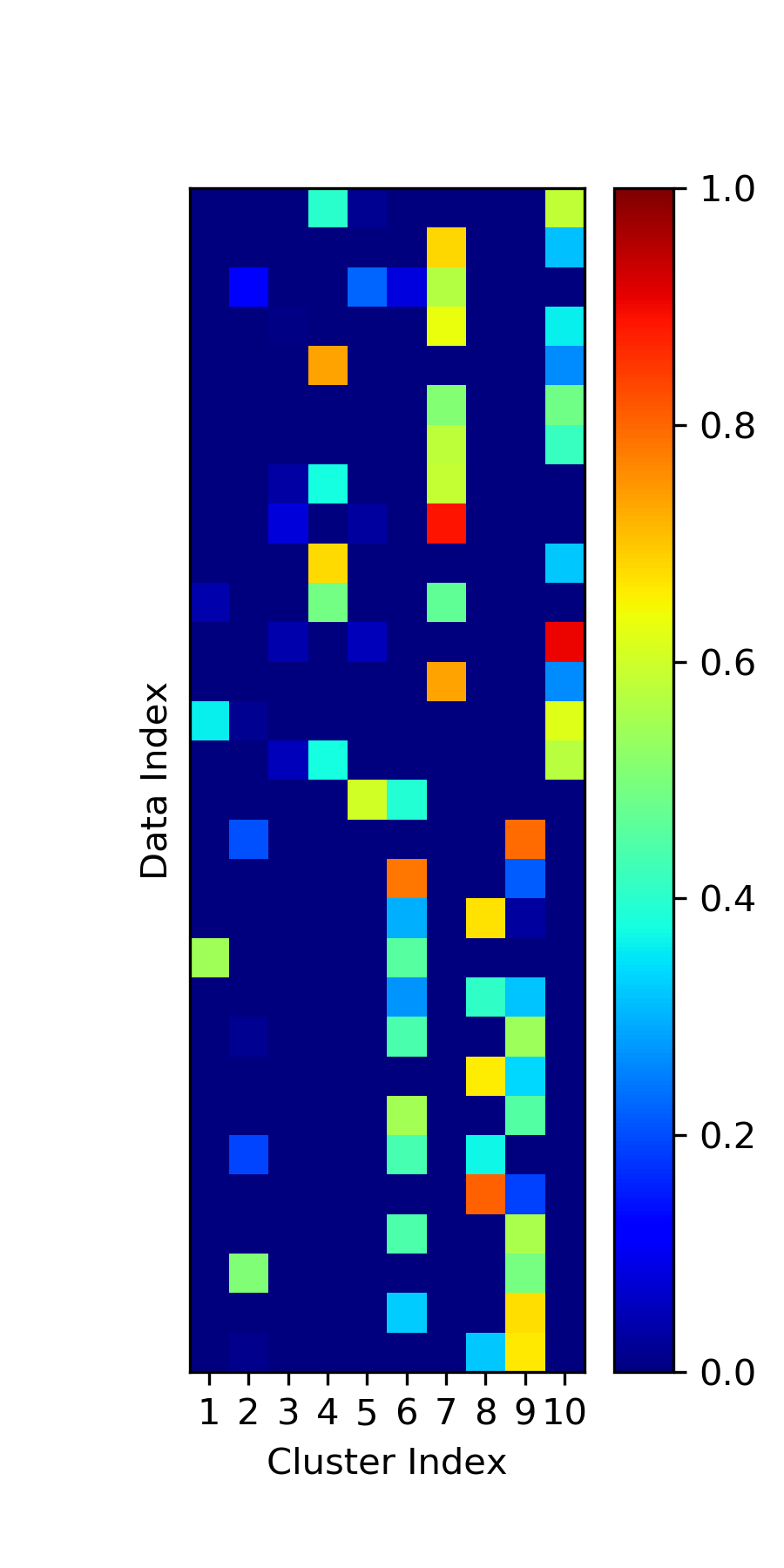}
      \end{subfigure}
      \caption{Estimated $W^{(l)}$ matrix using Dirichlet prior $\text{Dir}(0.1,\ldots,0.1)$ on each row.}
\end{subfigure}
 \caption{Group regularization on $W^{(l)}$ in an overfitted clustering model with $g=10$. \label{sim:rank_regularization}}
\end{figure}

Inspired by the group lasso variable selection in regression  \citep{yuan2006model,meier2008group},
 we use a group regularization to induce {\em column} sparsity in each $W^{(l)}$.
  \be
  R\big ( W^{(l)}\big )
    = \sum_{k=1}^g  \sqrt {\sum_{i=1}^n \big(\log \frac{w^{(l)}_{i,k}}{\epsilon}\big)^2_+}.
  \ee
   Since each $w^{(l)}_{i,k}\le 1$ and is already on a small scale, we first divide it by a close-to-zero and positive $\epsilon$, and take a logarithmic transform; those $w^{(l)}_{i,k}$ above and away from  $\epsilon\approx 0$ are penalized using $(x)_+=\max(x,0)$; those below $\epsilon$ are not penalized because they are negligibly small (in this article we use $\epsilon =10^{-3}$ as the threshold). To achieve a group regularization, a $2$-norm is used on each column; and $R\big ( W^{(l)}\big)$ is multiplied with $n$ so that it grows in the same order as $KL(P\| S)$ (which contains $O(n^2)$ terms). This regularization shows good empirical performance, as it recovers the true number of clusters in all of our simulation studies. Figure~\ref{sim:rank_regularization}(a) shows the estimated $W^{(l)}$ in the previous simulation.

We also consider some classical shrinkage prior on the simplex, such as the Dirichlet prior with the concentration parameter smaller than $1$. However, a drawback is that the shrinkage is applied independently on multiple simplex vectors, and there is no control on the {\em joint} distribution of all rows of $W^{(l)}$. As a result, many spurious small clusters appear even though each row is sparse [Figure~\ref{sim:rank_regularization}(b)]. We expect more advanced models such as hierarchical Dirichlet mixture \citep{teh2005sharing,zhou2014beta,ohama2017model} might solve this problem as well; we use the group regularization for computational convenience.

On the second regularization $T(\lambda_l)$, since there is only one simplex vector $(\lambda_1,\ldots, \lambda_d)$, it is easier to handle compared to $W^{(l)}$. We apply Dirichlet prior $\text{Dir}(\alpha_\lambda,\ldots,\alpha_\lambda)$, equivalently,
\be
T(\lambda_l) = (1-\alpha_\lambda)\log(\lambda_l),
\ee
and we use $\alpha_\lambda=1/d$ as a common choice for approximating the infinite mixture \citep{rasmussen2000infinite}.

With those two regularizations, we can choose $d$ and $g$ as large as possible (if the ground truth of the cluster number is not known). For example, we can choose the maximal $d$ and $g$ according to the computing budget, such as the memory limit.

\subsection{Producing Consensus via Combining Views}
In our model, the views with different $x^{(v)}$'s have distinct clustering patterns; on the other hand, sometimes there is still an interest to combine the information from those views together to form a `consensus'.

\begin{figure}[H]
 \begin{subfigure}[t]{0.45\textwidth}
 \centering
       \includegraphics[width=1\linewidth]{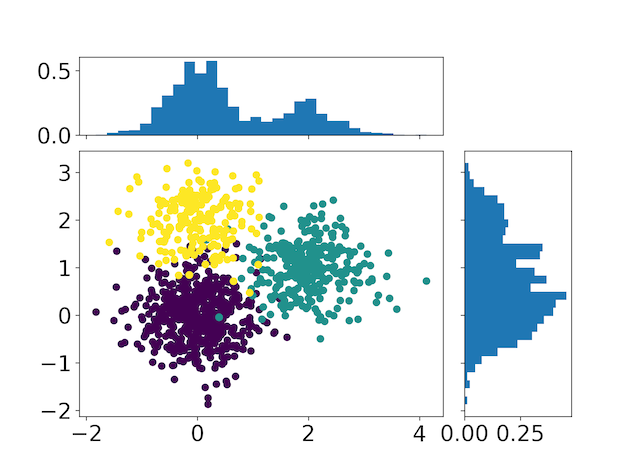}
             \caption{The first two views (each view is in $\mathbb{R}$), each corresponds to different clustering structure.}
 \end{subfigure}
  \rulesep
   \begin{subfigure}[t]{0.45\textwidth}
 \centering
       \includegraphics[width=1\linewidth]{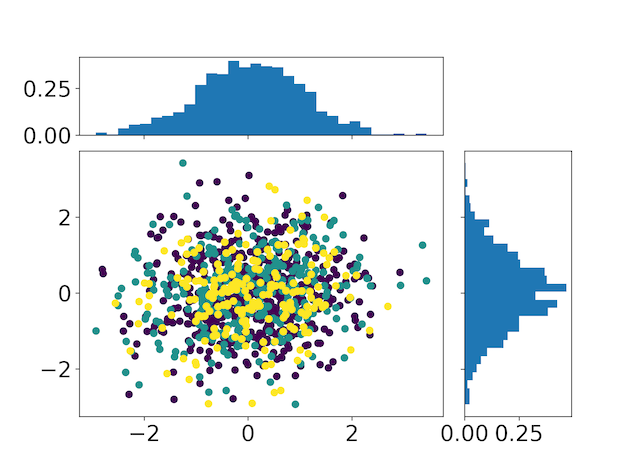}
             \caption{Two noisy views, neither contains any clustering structure.}
 \end{subfigure}
 \begin{subfigure}[t]{0.32\textwidth}
 \centering
       \includegraphics[width=1\linewidth]{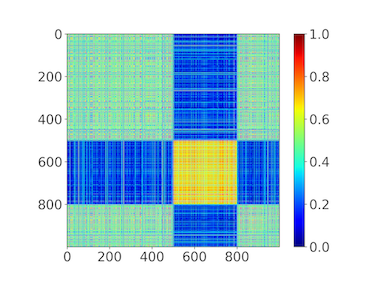}
             \caption{Estimated $P^{(1)}$ based on the first view.}
 \end{subfigure}
  \rulesep
\begin{subfigure}[t]{0.32\textwidth}
 \centering
       \includegraphics[width=1\linewidth]{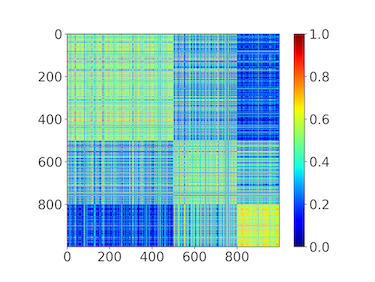}
             \caption{Estimated $P^{(2)}$ based on the second view.}
 \end{subfigure}
 \rulesep
 \begin{subfigure}[t]{0.32\textwidth}
 \centering
       \includegraphics[width=1\linewidth]{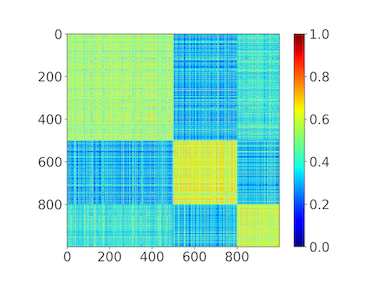}
             \caption{Consensus formed by combining the first two views, while ignoring the other ones.}
 \end{subfigure}
  \caption{Illustration on how to combine different views to form a consensus (each view in $\mathbb{R}$: the estimated $P^{(v)}$'s from the first two views (two marginal histograms of panel a) are taken, while the ones with no clustering structures (panel b) are excluded, producing an average as the consensus co-assignment probability matrix (panel e). \label{fig:combining_subspaces}}
 \end{figure}

Since a convex combination of $P^{(v)}$'s is still positive semi-definite, we consider the weighted average as the `consensus' co-assignment probability
\bel
\label{eq:concensus}
\bar P = \sum_{v=1}^V \frac{u^{(v)}  P^{(v)}}{\sum_{v=1}^V u^{(v)} }.
\eel
with the weight $u^{(v)}\ge 0$, and $\sum_{v=1}^V u^{(v)}>0$. In this article, we take a simple strategy for choosing $u^{(v)}$: for a view, if its most probable $\hat x^{(v)}$ (which can be computed from \eqref{eq:most_probable_param} in Section 3) is equal to $l$ and $W^{(l)}$ is a matrix with one column filled by $1$'s and others by $0$'s, then there is no clustering pattern in this view; hence, we set its $u^{(v)}$ to zero. For the other views, we set equal weights $u^{(v)}=1$.

This consensus is closely related to the variable-selection based clustering \citep{witten2010framework}. Indeed, the latter is equivalent to directly using $d=2$ latent parameterizations (one of them having no clustering).

To illustrate the consensus, we generate data with $10$ views, each in $\mathbb{R}$: each of the first two views contains more than one clusters [Figure~\ref{fig:combining_subspaces}(a), generated from $N(0,1),N(2,1)$ and $N(0,1),N(1,1),N(2,1)$], respectively; while the views $3-10$ have no clustering structure [Figure~\ref{fig:combining_subspaces}(b), all generated from $N(0,1)$]. In the estimation of the LSP model, $P^{(3)},\ldots,P^{(10)}$ are linked to an $W^{(l)}$ without clustering structure. Using \eqref{eq:concensus}, the consensus combines  $P^{(1)}$ and $P^{(2)}$  and shows that there are three clusters.

\section{Computation}
As $x^{(v)}$ is a latent variable, we use the Expectation-Maximization (EM) algorithm. Letting $\eta^{(v)}_l=\mathbb{E} 1(x^{(v)}=l)$, in the E step, we update
\bel\label{eq:E_step}
\eta^{(v)}_l =  \frac{\lambda_l  \exp\big[ -\sum_{j<i}  KL \big(p^{*(l)}_{i,j}\|s^{(v)}_{i,j}\big)  \big] }
{\sum_{l'=1}^d\lambda_{l'}   \exp\big[- \sum_{j<i}  KL \big(p^{*(l')}_{i,j}\|s^{(v)}_{i,j}\big)
 \big]},
\eel
where $p^{*(l)}_{i,j}= (W^{(l)} W^{(l)\rm T})_{i,j}$.

In the M step, we minimize the expected loss function over the parameter $W^{(l)}$, using the ADAM gradient descent algorithm  \citep{kingma2014adam}:
\bel\label{eq:M_step}
\mathbb{E}_{x^{(1)},\ldots,x^{(V)}} RegLoss =
& \sum_{v=1}^V \sum_{l=1}^d \eta^{(v)}_l
 \sum_{j<i}  \bigg [{p^{*(l)}_{i,j}}\log \frac{p^{*(l)}_{i,j}}{s^{(v)}_{i,j}}+
{(1-p^{*(l)}_{i,j})}\log \frac{1-p^{*(l)}_{i,j}}{1-s^{(v)}_{i,j}}\bigg] \\
&  +   n\sum_{l=1}^d R\big ( W^{(l)}\big )    + \sum_{l=1}^d T(\lambda_l),
\eel
and set $\lambda_l$ to its mode
\be
\hat \lambda_l \propto \max(0, \alpha_\lambda -1 + \sum_{v=1}^V \eta^{(v)}_l  ) \text{ for } l=1,\ldots,d \text{ such that } \sum_{l=1}^d \hat \lambda_l =1.
\ee

The vector $(\eta^{(v)}_1, \ldots,\eta^{(v)}_d)$ gives the scores on how likely the $v$th view is generated from each parameterization. As a point estimate for $x^{(v)}$, the most probable one is
\bel\label{eq:most_probable_param}
\hat x^{(v)} = \underset{l\in \{1,\ldots,d\}}{\arg\max} \; \eta^{(v)}_l .
\eel
Similarly, we have the point-wise optimal $\tilde c^{(v)}_i=\arg\max_k w^{(\hat x^{(v)})}_{i,k}$. We can use those two quantities to determine the effective numbers of parameterizations and clusters: $\hat d$ as the number of unique $\hat x^{(v)}$'s for $v=1,\ldots, V$, and $\hat g^{(v)}$ as the number of unique $\tilde c^{(v)}_i$'s for $i=1,\ldots, n$.

On the other hand, as shown by \cite{wade2018bayesian}, the point-wise clustering estimate $\tilde c^{(v)}_i$ is not necessarily optimal for the overall clustering. Instead, we use the estimated $P^{(\hat x^{(v)})}$ and $\hat g^{(v)}$ as the input matrix and cluster number in the spectral clustering, to produce a joint point estimate  $(\hat c^{(v)}_{1}, \ldots, \hat c^{(v)}_{n} )$. As shown in the data experiments, this results in much more accurate clustering than using $S^{(v)}$ directly in the spectral clustering.

\subsection{Scalability and Initialization}
In our optimization algorithm, the M step is the most computationally intensive one, since we need several gradient descents in each EM iteration. Fortunately, we can substantially reduce its computing complexity and make it scalable to a very large $V$.

Before using gradient descent, we first compute two matrices, with their $(i,j)$th elements
\be
\kappa_{i,j}=
-\sum_{v=1}^{V}\eta^{(v)}_l\log \frac{s^{(v)}_{i,j}}{1-s^{(v)}_{i,j}}, \qquad \gamma_{i,j}=\sum_{v=1}^{V}\eta^{(v)}_l
.
\ee
By changing the order of summation, the expected loss in \eqref{eq:M_step} becomes
\be
\sum_{l=1}^d&   \sum_{j<i}  \bigg \{ \kappa_{i,j}p^{*(l)}_{i,j} +
\gamma_{i,j}  \bigg[ p^{*(l)}_{i,j} \log \frac{p^{*(l)}_{i,j}}
{1- p^{*(l)}_{i,j} } +   \log (1-   p^{*(l)}_{i,j} )  \bigg ]  \bigg \} \\
&   +   n\sum_{l=1}^d R\big ( W^{(l)}\big )    + \sum_{l=1}^d T(\lambda_l)+ C,
\ee
where $C = - \sum_{v=1}^V \sum_{l=1}^d \eta^{(v)}_l \sum_{j<i}   \log(1-s^{(v)}_{i,j})$ is a constant free  from $W^{(l)}$; hence it can be ignored during the M step.
Notice that this alternative form reduces the computational complexity from  $O(Vn^2d)$ to $O(n^2d)$ for each gradient descent.

Similar to the conventional mixture models, when starting the
EM algorithm, it is crucial to have good initial values for the parameters. Therefore, we now develop an initialization strategy. Note that if ignoring the low-rank constraint in $p^{*(l)}_{i,j}$'s, the loss for those $x^{(v)}=l$, $\sum_{v: x^{(v)}=l} KL \big(p^{*(l)}_{i,j}\|s^{(v)}_{i,j}\big)$
is minimized at
\be
\log \frac{\widehat p^{*(l)}_{i,j}}{1- \widehat p^{*(l)}_{i,j}} = \big[{\sum_v 1(x^{(v)}=l)}\big ]^{-1}\sum_{v: x^{(v)}=l}\log \frac{s^{(v)}_{i,j}}{1-s^{(v)}_{i,j}},
\ee
which is the group mean of the log odds. Therefore, we first use a simple K-means (with K set to $d$) on $V$ matrices $\{\log[{s^{(v)}_{i,j}}/{(1-s^{(v)})}]\}_{i,j}$,
putting them into $d$ groups and treating the K-means labels as the initial estimates for $\hat x^{(v)}$'s. Then setting $\eta^{(v)}_l=1(\hat x^{(v)}=l)$ and $\lambda_l=1/d$, we run the M step to obtain the initial values for $W^{(l)}$'s.

We track the expected loss $\mathbb{E}_{\{x^{(v)}\}} RegLoss$ for convergence, and consider the algorithm as converged if the decrease in the expected loss is less than one percent over 100 iterations. Since the loss function is non-convex, we run the algorithm multiple times under random initializations with K-means++\citep{arthur2007k}. We choose the ones with the lowest loss as the final estimates.

\section{Theory}
In this section, we provide a theoretical justification for the LSP model, by establishing a link to the random partition distribution using pairwise information/distances \citep{blei2011distance,dahl2017random}.

We first briefly review the idea of random partition distribution.
Given a matrix of co-assignment probabilities $ P=\{ p_{i,j}\}_{i,j}$, with $p_{i,j}\in(0,1)$, we can sample a cluster graph. Starting with an initial set containing one index $\mathcal N_0=\{ i \}$ (with $i$ randomly chosen), each time, we draw another $j$ randomly from $\mathcal N \setminus \mathcal N_0$ and assign
\begin{equation}
\label{eq:seq_prob_distribution}
        \begin{aligned}
 &   \hat z_{i,j}^{(v)} = 1 ,& \text{ if }   \forall  \hat z^{(v)}_{i,i'}=1, \hat z^{(v)}_{j,i'}=1,\\
 &   \hat z_{i,j}^{(v)} = 0 , & \text{ if }  \forall \hat z^{(v)}_{i,i'}=0, \hat z^{(v)}_{j,i'}=1,\\
  &  \hat z_{i,j}^{(v)} = 0 ,  &\text{ if } \forall\hat   z^{(v)}_{i,i'}=1, \hat z^{(v)}_{j,i'}=0,\\
&        \hat z_{i,j}^{(v)}\sim \text{Bernoulli}( p_{i,j})  ,&  \text{
otherwise},
        \end{aligned}
\end{equation}
for $i,i'\in \mathcal N_0$ and $i\neq i'$. That is, sampling new Bernoulli if it is not determined by the pairwise constraints of a cluster graph. After updating $\hat z^{(v)}_{i,j}$, we add $j$ into $\mathcal N_0$ and go to the next loop.
Eventually, this forms an $n\times n$ binary matrix $\hat Z^{(v)}$. These procedures are associated with a random partition distribution, that we denote by $\hat Z^{(v)}\sim\phi$.

Now under the multi-view setting, let us focus on the sub-group of views with the same parameterization (that is, having equal $x^{(v)}$'s). Without loss of generality, we assume they have indices $v=1,\ldots,M$.

For each graph, we assume that there is a ground truth cluster graph $Z^{(v)}_0$. If it is known, we can compare it with the sampled $\hat Z^{(v)}$ and compute a loss function in $[0,1]$ (such as the $1$-minus normalized mutual information)
\begin{equation*}
  \begin{aligned}
    loss \left(Z_0^{(v)}, \hat Z^{(v)}\right).
  \end{aligned}
\end{equation*}
To assess the quality of $\phi$ in recovering the ground truth, theoretically, we would hope to take average over infinite samples of $\hat Z^{(v)}\sim \phi$, and then take  average over $M$ views,
\be
R(\Pi_M, \phi)= \frac{1}{M}
\sum_{v=1}^M
\mathbb{E}_{\hat Z^{(v)}
\sim
\phi}    loss \left(Z_0^{(v)}, \hat Z^{(v)}\right),
\ee
where $\Pi_M$ denotes the empirical distribution for $M$ views.
Taking one step further, suppose $Z^{(v)}_0 \stackrel{iid}{\sim} \Pi_0$, we can define the generalization risk
\be
R(\Pi_0, \phi)=
  \mathbb{E}_{Z_0^{(v)}\sim \Pi_0}   \mathbb{E}_{\hat Z^{(v)} \sim
\phi}  loss \left(Z_0^{(v)}, \hat Z^{(v)}\right).
\ee

Since $\Pi_0$ is not fully known, we cannot directly minimize $R(\Pi_0, \phi)$; however,  we can use some assumptions on $\Pi_0$ (such as having at most $g$ clusters) and use $R(\Pi_M, \phi)$ as an approximate. Therefore, it is imperative to optimize approximation.  We have the following bound based on the Probably Approximately Correct (PAC)-Bayes theory \citep{seldin2010pac,guedj2019primer}.



\begin{theorem}
For $M\in  [2,\infty)$, if $(1/{M})
\sum_{v=1}^M loss \left(Z_0^{(v)}, \hat Z^{(v)}\right)\in(0,1)$, with probability greater than $1-\delta$ based on $\Pi_0$ (the true distribution for $Z^{(v)}_0$),
\begin{equation*}
  \begin{aligned}
    KL &\big[ R(\Pi_0, \phi) \;||\; R(\Pi_M, \phi)\big] \\
   & \le  (1/M) \bigg\{
   \sum_{v=1}^M\sum_{j<i}  KL( p_{i,j}|| s^{(v)}_{i,j})/M + \log \bigg[ \exp(\frac{1}{12M}) \sqrt{\frac{\pi M}{2}} + 2 \bigg] - \log \delta \bigg\}
      ,
\end{aligned}
\end{equation*}
where  $ p_{i,j}$ is absolutely continuous with respect to  $s^{(v)}_{i,j}$.
\end{theorem}
Combining the $KL$ terms on the right hand side over the different parameterizations, we  obtain the $KL(P\| S)$ function in our LSP model.

Therefore, estimating the LSP model can be considered as a procedure to optimize the multi-view random partition distribution $\phi$, in terms of improving the finite-view performance. Specifically, minimizing the difference between $R(\Pi_0, \phi)$ and $R(\Pi_M, \phi)$ reduces the chance of overfitting, and is known as `reducing generalization error' in the PAC-Bayes literature \citep{seldin2010pac}. Compared to the canonical random partition distribution, we also gain in the computation since we do not have to sample $\hat Z^{(v)}$.


\section{Data Experiments}

\subsection{Single View Simulations}
\begin{figure}[H]
 \begin{subfigure}[t]{0.3\textwidth}
 \centering
       \includegraphics[width=1\linewidth]{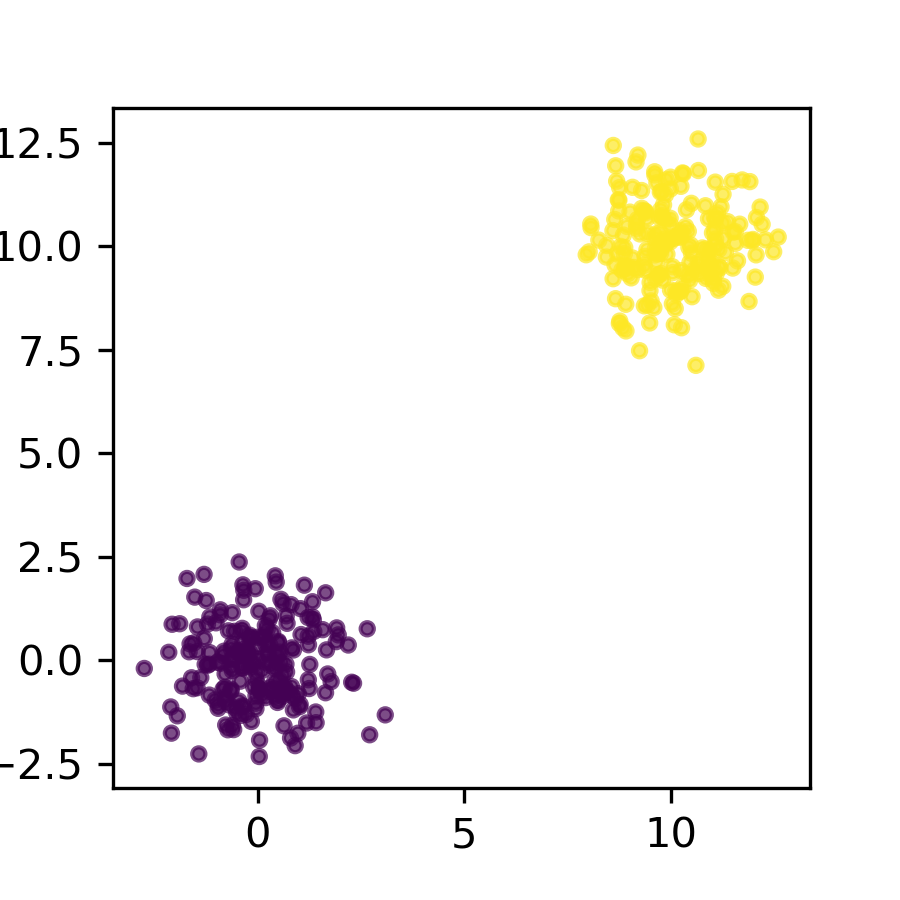}
             \caption{Gaussian: \\ $N[(0,0)', I]$ \\ and $N[(10,10)',I]$.}
 \end{subfigure}
 \rulesep
\begin{subfigure}[t]{0.3\textwidth}
 \centering
       \includegraphics[width=1\linewidth]{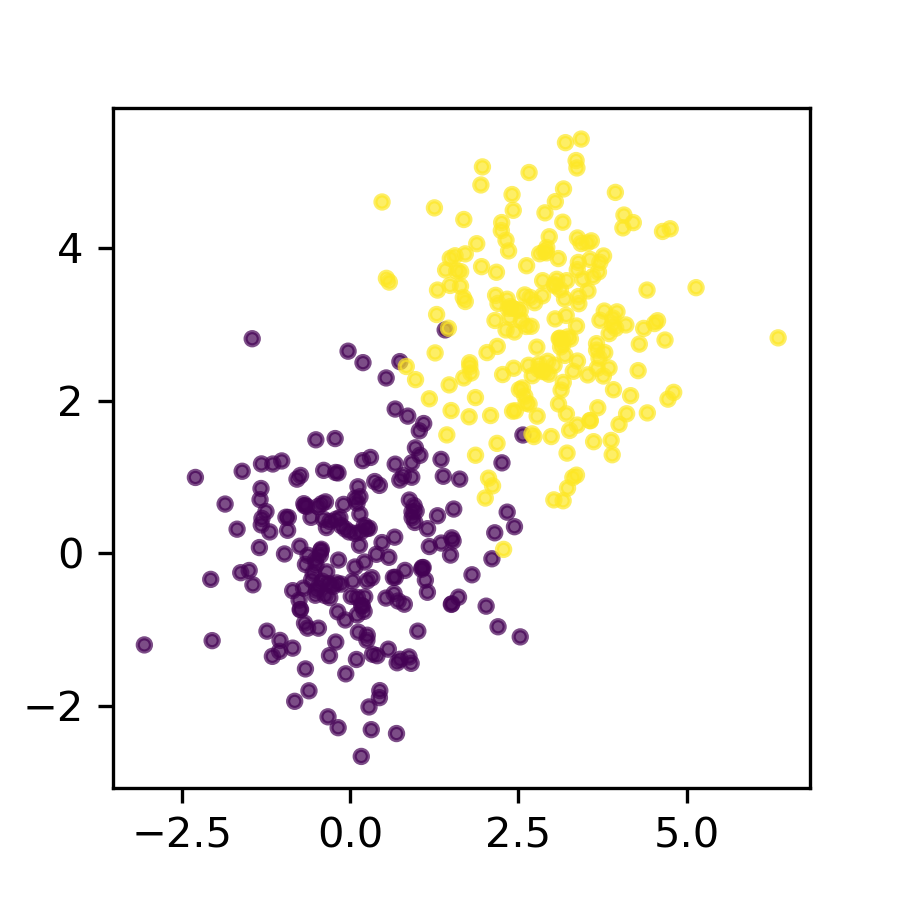}
             \caption{Gaussian: \\ $N[(0,0)', I]$ \\and  $N[(3,3)',I]$.}
 \end{subfigure}
 \rulesep
 \begin{subfigure}[t]{0.3\textwidth}
 \centering
       \includegraphics[width=1\linewidth]{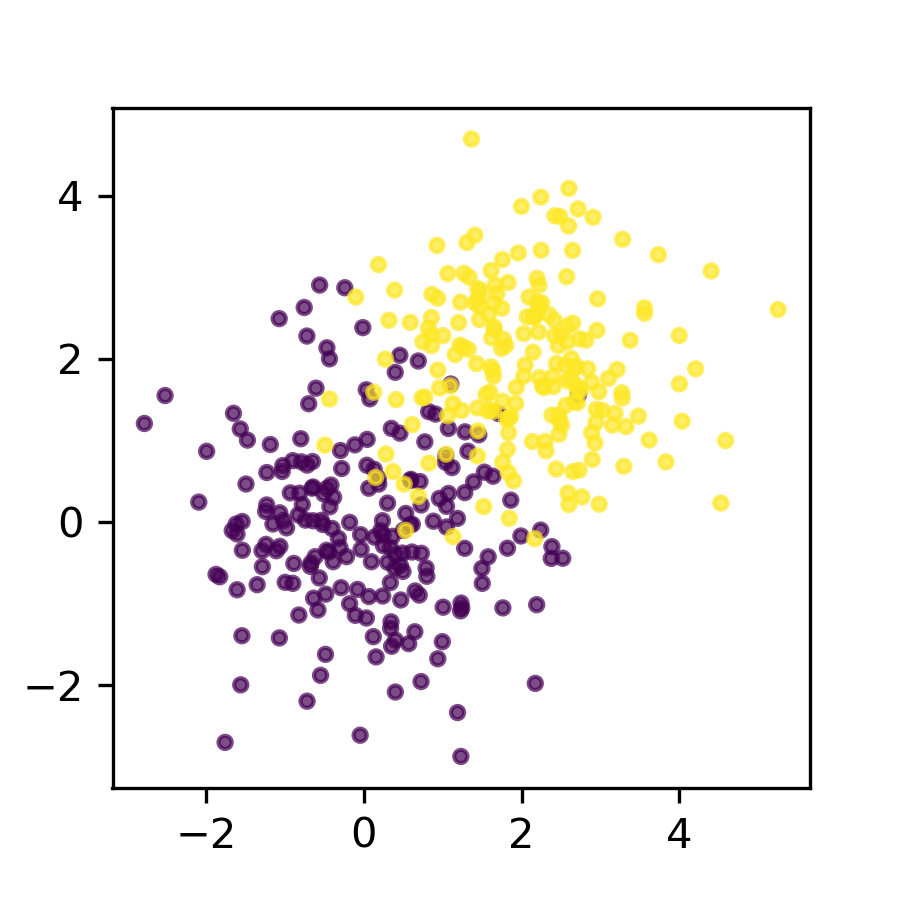}
             \caption{Gaussian: \\
               $N[(0,0)', I]$ \\and  $N[(2,2)',I]$.}
 \end{subfigure}\\
  \begin{subfigure}[t]{0.3\textwidth}
 \centering
       \includegraphics[width=1\linewidth]{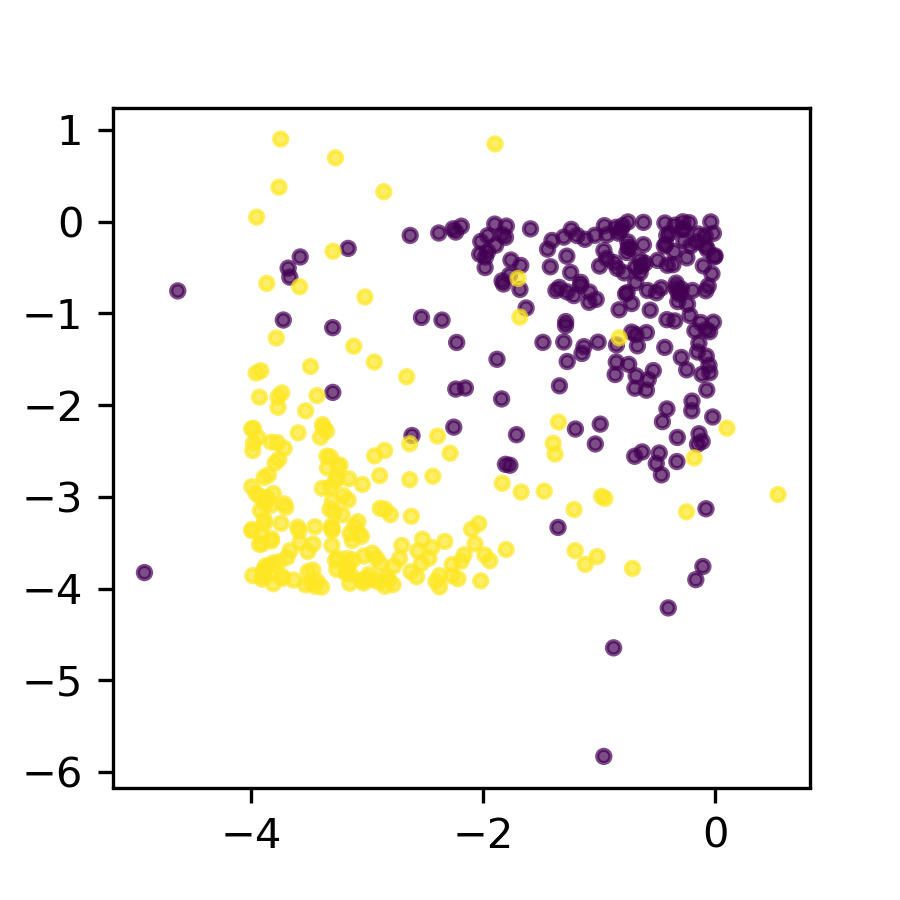}
             \caption{ \small Shifted exponential:\\ $[\text{Exp}(1)-4, \text{Exp}(1)-4]$
         \\            and $[-\text{Exp}(1), -\text{Exp}(1)]$}
 \end{subfigure}
 \rulesep
 \begin{subfigure}[t]{0.3\textwidth}
 \centering
       \includegraphics[width=1\linewidth]{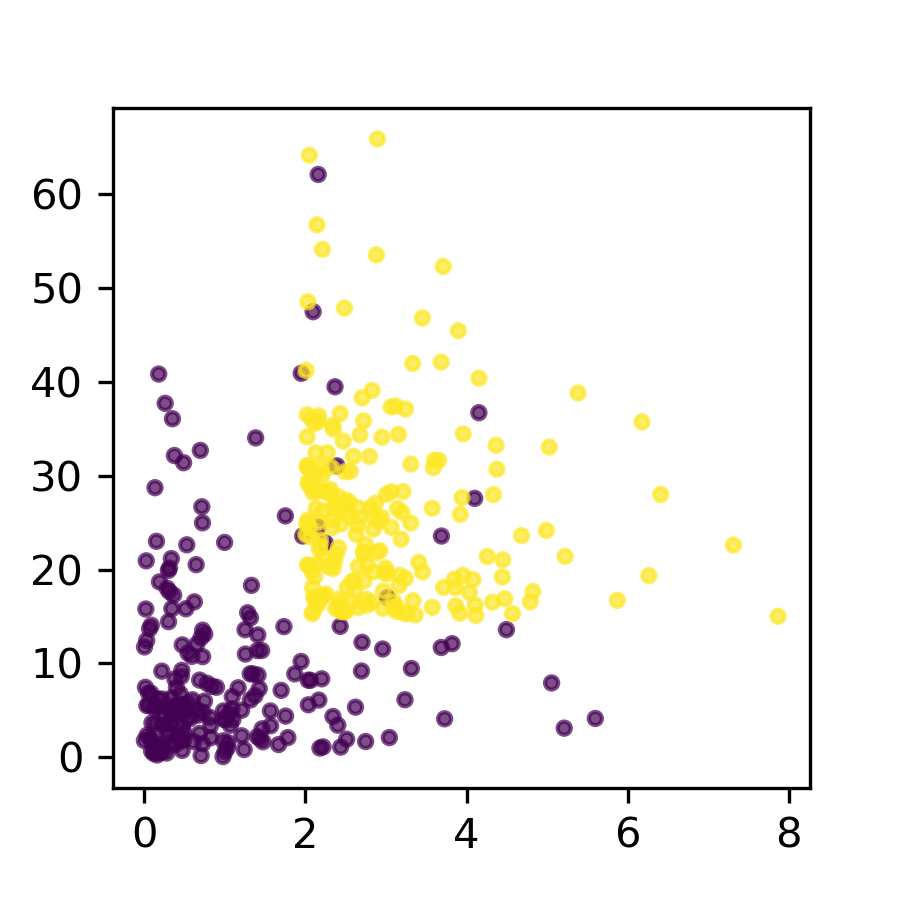}
             \caption{ \small Shifted exponential:\\ $[\text{Exp}(1), \text{Exp}(10)]$ \\            and $[\text{Exp}(1)+2, \text{Exp}(10)+15]$}
 \end{subfigure}
 \rulesep
\begin{subfigure}[t]{0.3\textwidth}
 \centering
       \includegraphics[width=1\linewidth]{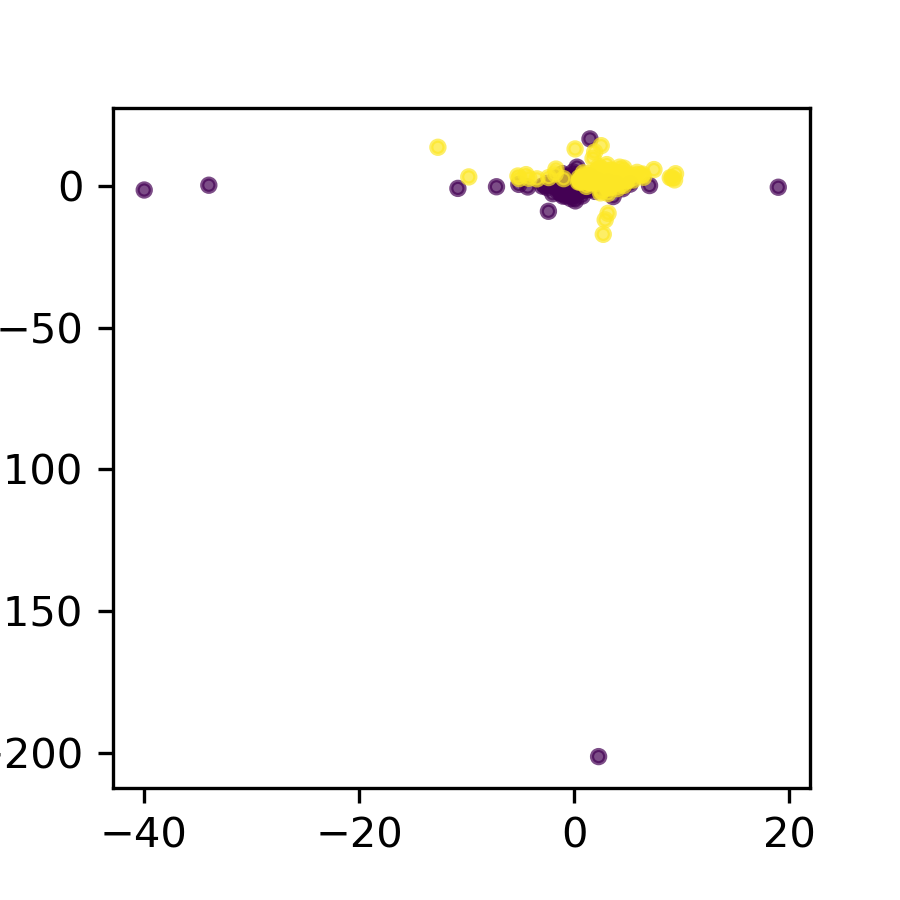}
             \caption{ \small Shifted Cauchy:\\ $[t_1,t_1]$
\\          and  $[t_1+3,t_1+3]$}
 \end{subfigure}
 \caption{Generated data in the single view clustering experiments (the view is in $\mathbb{R}^2$). In each setting, two clusters are simulated, with the color representing the ground truth labels.  \label{fig:uncertainty_quantification}}
 \end{figure}

 Since most clustering approaches are based on a single view, we first compare our model with them using simulations. For a clear visualization, we generate data from the two-component mixture distribution in a single view $y_i \in \mathbb{R}^2$, with $n=400$.  Figure~\ref{fig:uncertainty_quantification}(a-f) plots the generated data under $6$ different settings.

For comparison, we test $9$ other clustering algorithms, provided by the Scikit-Learn Cluster package. We use the Normalized Mutual Information (NMI) as a benchmark score. NMI measures the accuracy of the estimated cluster labels with respect to the ground truth labels and is invariant to label-switching. The result is listed in Table~\ref{tb:single_view_exp}. In the Gaussian cases (a-c), the Gaussian mixture as the true model has the best performance. When the symmetric and Gaussian-tail assumptions are violated (d-f), more recent methods such as spectral clustering start to show their advantage. The performance of the LSP model (with $g=2$ and $d=1$) is very close to the spectral clustering; while in (f) with the heavy-tailed distribution creating many `outliers', the spectral clustering fails completely due to the large noise in the raw similarity matrix, and LSP does not have this issue thanks to the low-rank smoothing.

 \begin{table}[H]
     \centering
     \begin{tabular}{ l | l l l l l l}
         \hline
         &(a)     & (b)        & (c)        & (d)   & (e)        & (f) \\
         \hline
         K-means                               &\bf{1}  & 0.89       & 0.68       & 0.56  & 0.25       & 0\\
         Affinity propagation &0.57    & 0.37       & 0.30       &0.31   & 0.30       & 0.23 \\
         Mean-shift                            &\bf{1}  & 0.89       & \bf{0.70}  &0.56   & 0.40       & 0.05\\
         Agglomerative clustering              &\bf{1}  & 0.77       & 0.62       &0.54   & 0.17       & 0\\
         DBSCAN                       &0.92    & 0          & 0.01       &0      & 0.43       & 0\\
         OPTICS                       &0.23    & 0.20       & 0.17       &0.21   & 0.19       & 0.17\\
         Gaussian mixture                     &\bf{1}  & \bf{0.91}  & \bf{0.70}  & 0.58  & 0.37       & 0\\
         Birch                                 &0.99    & 0.70       & 0.49       & 0.41  & 0.27       & 0.01\\
         Spectral clustering                   &\bf{1}  & 0.89       & \bf{0.70}  & {\bf 0.59}  & 0.55       & 0.0\\
         LSP                                   &\bf{1}  & 0.89       & \bf{0.70}  & {\bf 0.59}  & \bf{0.58}  & \bf{0.42}\\
         \hline
     \end{tabular}
     \caption{ Normalized mutual information showing the accuracy of single view clustering, using the data simulated in Figure~\ref{fig:uncertainty_quantification}. \label{tb:single_view_exp}}
 \end{table}
For uncertainty quantification, we compute the co-assignment probability using the ground truth distribution and compare it with the estimated $P$ from the LSP model. The raw $S$ and estimated $P$ are provided in the appendix. The median absolute deviations (MAD) between $P$ and the oracle are  $0.03$, $0.07$, $0.08$, $0.04$, $0.05$ and $0.06$. In addition, to assess the limitation of the LSP model (and similarity-based algorithms in general), we modify (c) and make the two clusters closer. When two clusters are generated from $N[(0,0)', I]$ and  $N[(1,1)',I]$, the LSP model could only discover one large cluster. This is not surprising since heavily overlapped clusters can be alternatively taken as one cluster; in such cases, mixture models with stronger assumptions would work better, such as the ones with a parametric density for each cluster.
\subsection{Multi View Experiments}
\subsubsection{Scaling to a large number of views}
\begin{figure}[H]
\begin{subfigure}[t]{1\textwidth}
 \begin{subfigure}[t]{0.19\textwidth}
 \centering
       \includegraphics[width=1\linewidth]{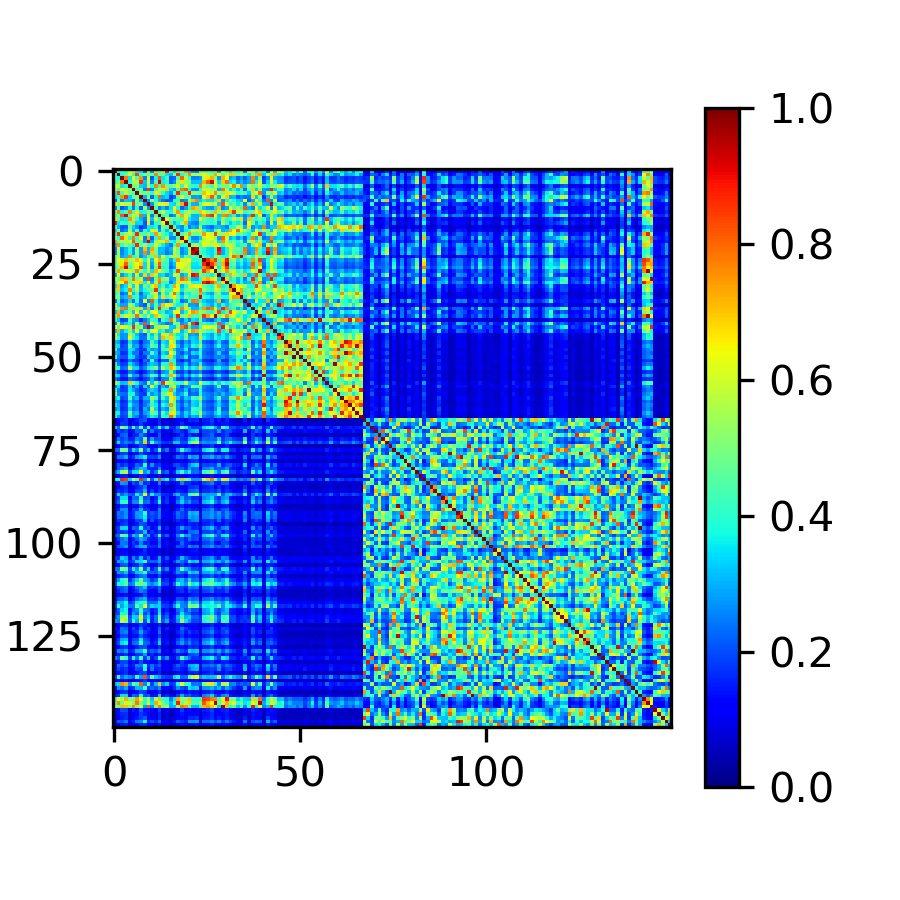}
 \end{subfigure}
\begin{subfigure}[t]{0.19\textwidth}
 \centering
       \includegraphics[width=1\linewidth]{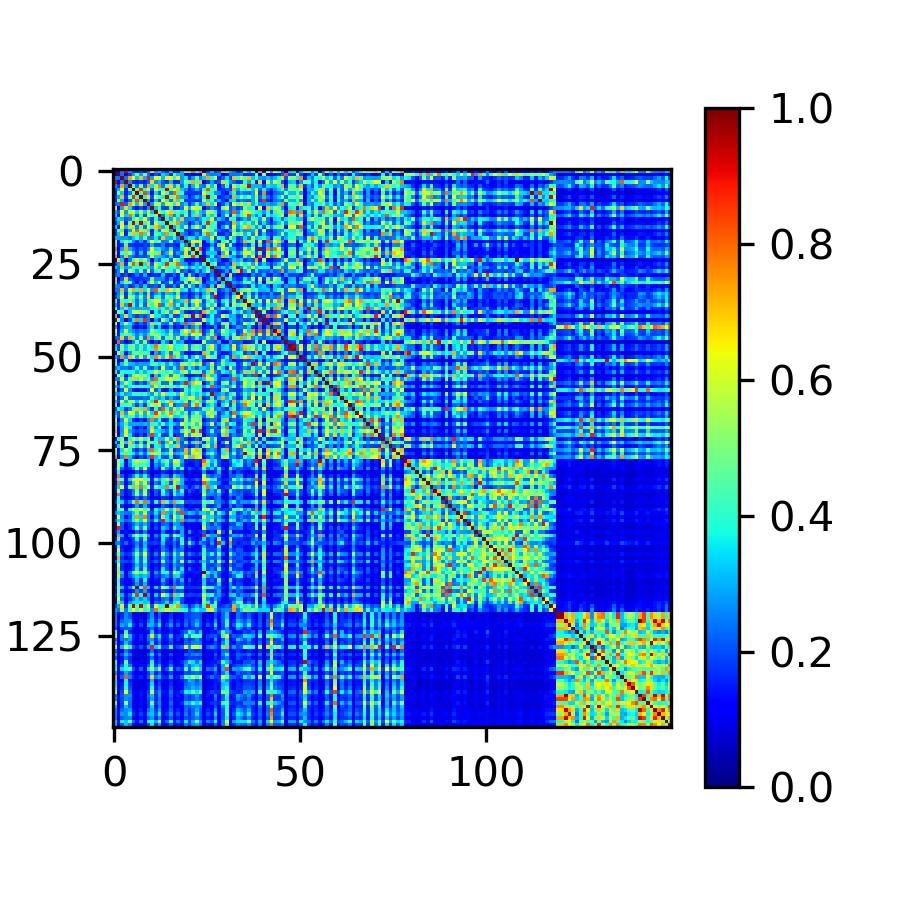}
 \end{subfigure}
 \begin{subfigure}[t]{0.19\textwidth}
 \centering
       \includegraphics[width=1\linewidth]{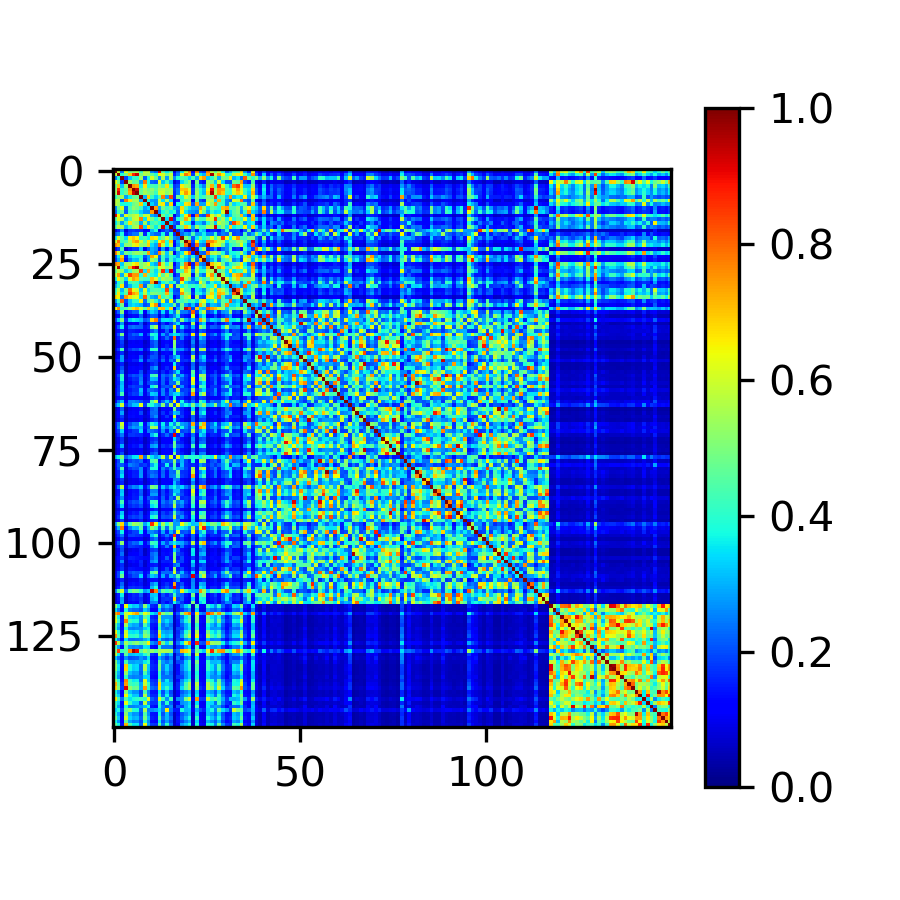}
 \end{subfigure}
  \begin{subfigure}[t]{0.19\textwidth}
 \centering
       \includegraphics[width=1\linewidth]{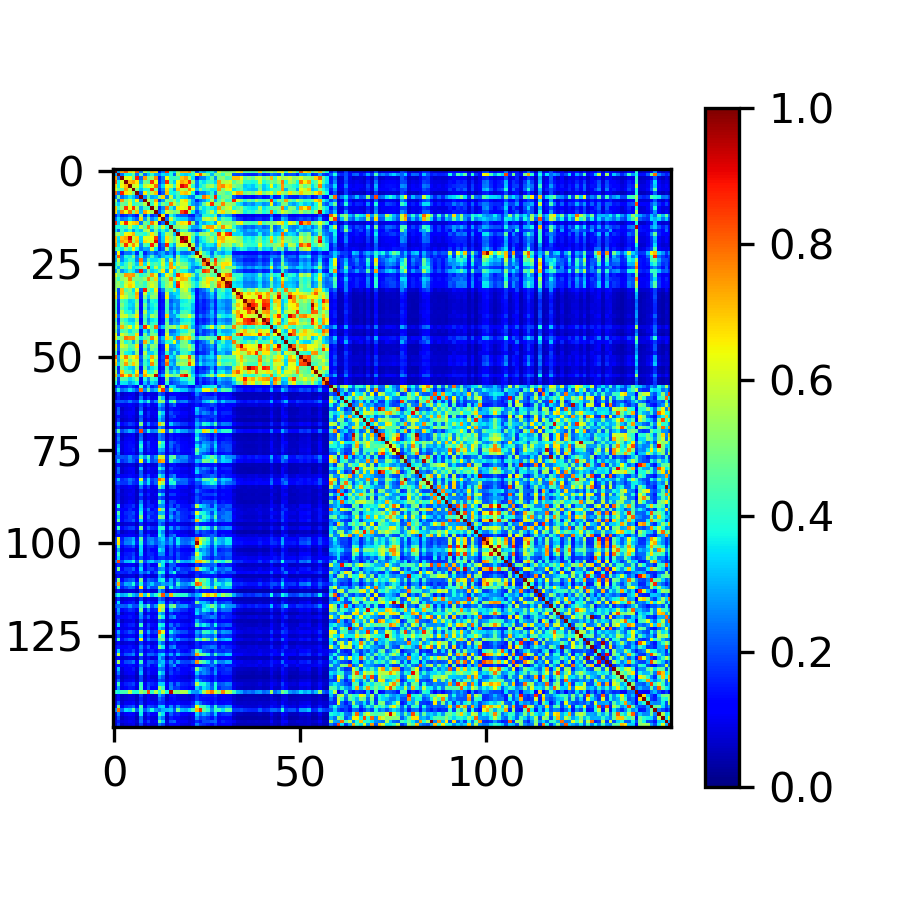}
 \end{subfigure}
  \begin{subfigure}[t]{0.19\textwidth}
 \centering
       \includegraphics[width=1\linewidth]{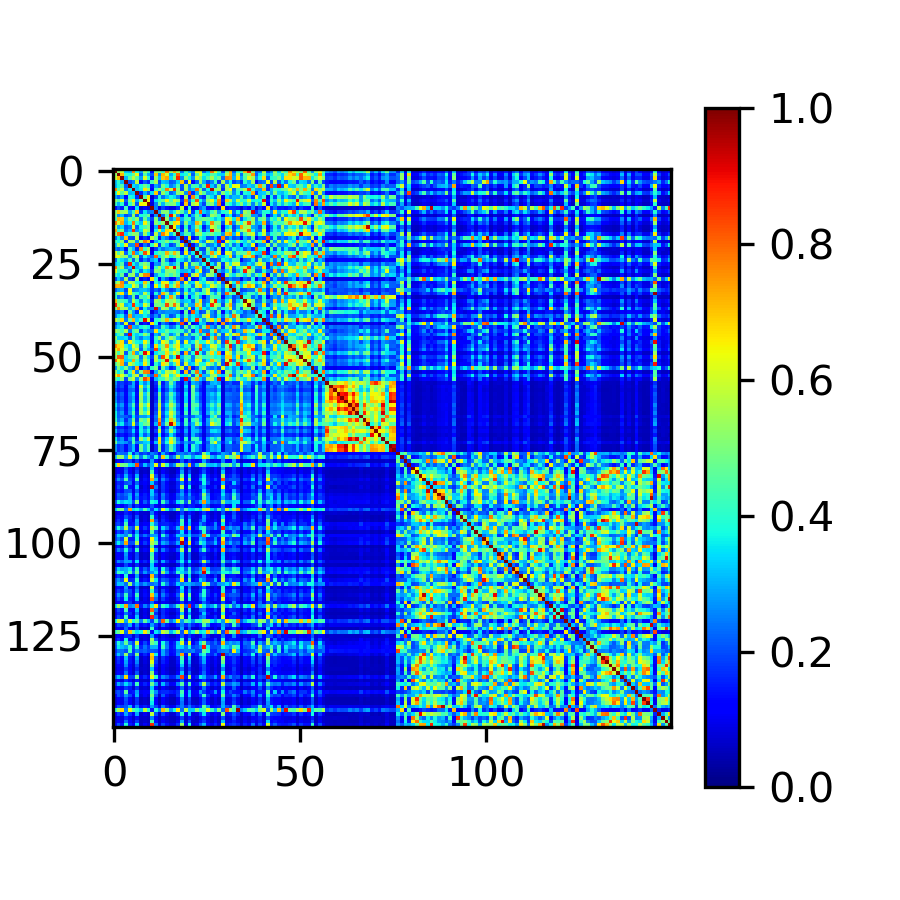}
 \end{subfigure}
    \caption{Five similarity matrices corresponding to different clustering patterns.}
 \end{subfigure}
 \begin{subfigure}[t]{1\textwidth}
 \begin{subfigure}[t]{0.19\textwidth}
 \centering
       \includegraphics[width=1\linewidth]{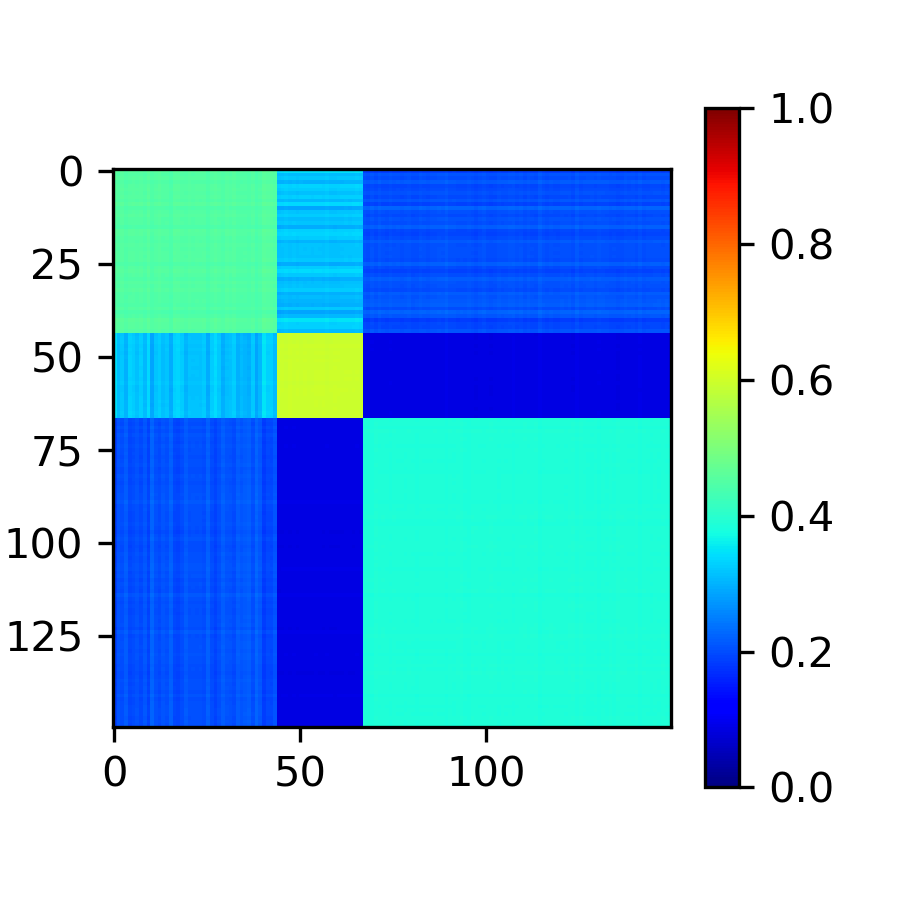}
 \end{subfigure}
\begin{subfigure}[t]{0.19\textwidth}
 \centering
       \includegraphics[width=1\linewidth]{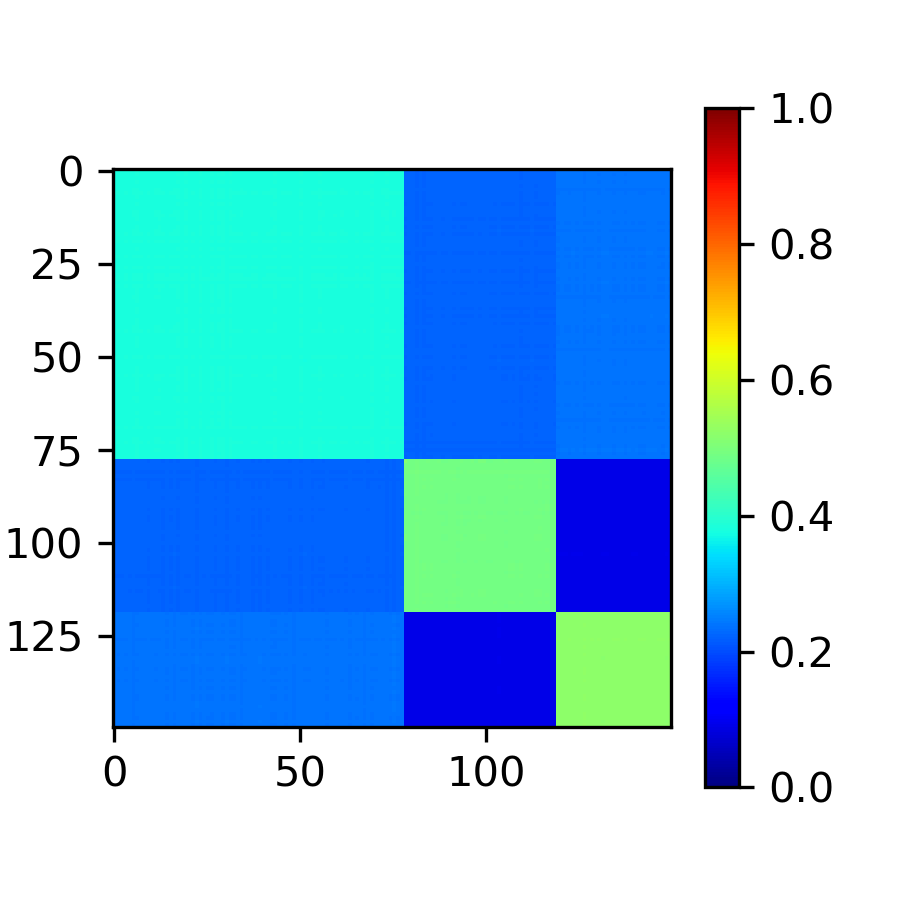}
 \end{subfigure}
 \begin{subfigure}[t]{0.19\textwidth}
 \centering
       \includegraphics[width=1\linewidth]{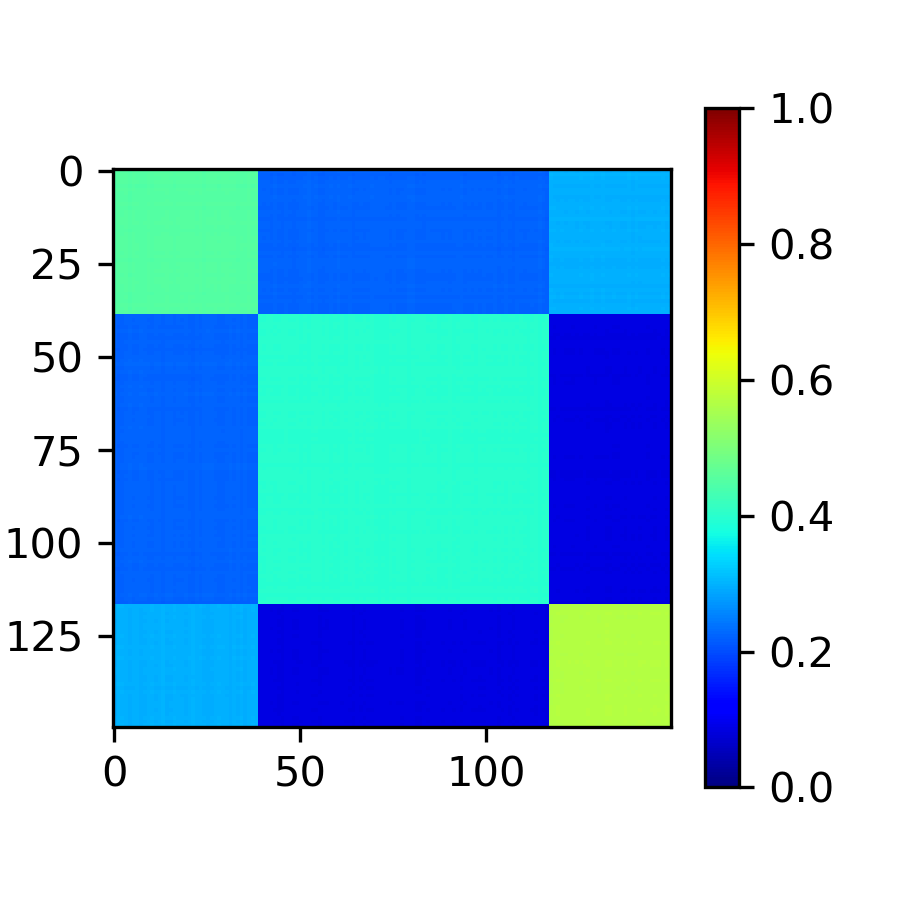}
 \end{subfigure}
  \begin{subfigure}[t]{0.19\textwidth}
 \centering
       \includegraphics[width=1\linewidth]{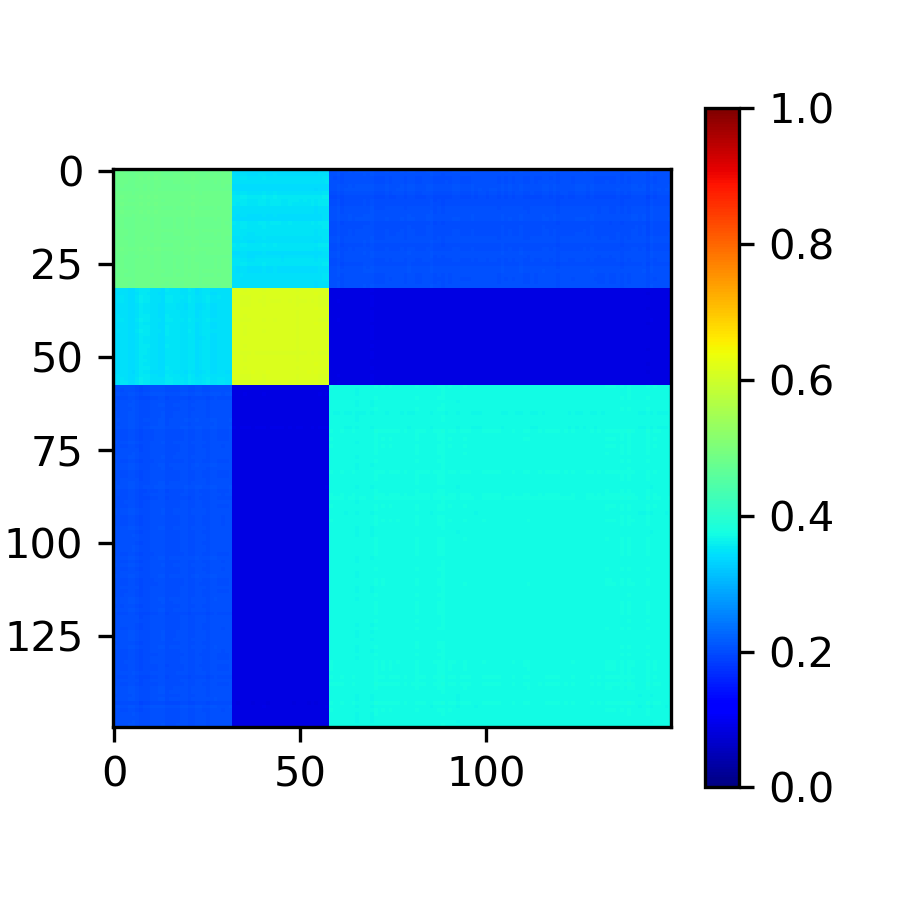}
 \end{subfigure}
  \begin{subfigure}[t]{0.19\textwidth}
 \centering
       \includegraphics[width=1\linewidth]{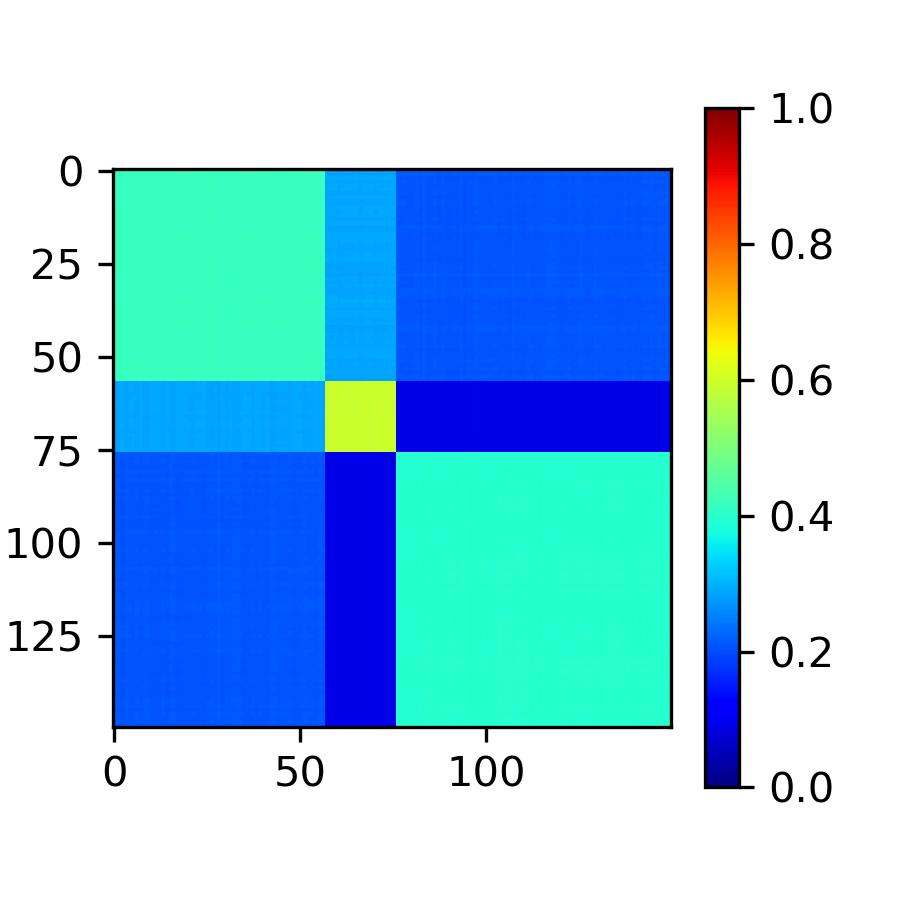}
 \end{subfigure}
    \caption{Estimated parameterization $W^{(l)}W^{(l) \rm T}$.}
 \end{subfigure}
\begin{subfigure}[t]{1\textwidth}\centering
   \begin{subfigure}[t]{1\textwidth}
 \centering
       \includegraphics[width=.4\linewidth]{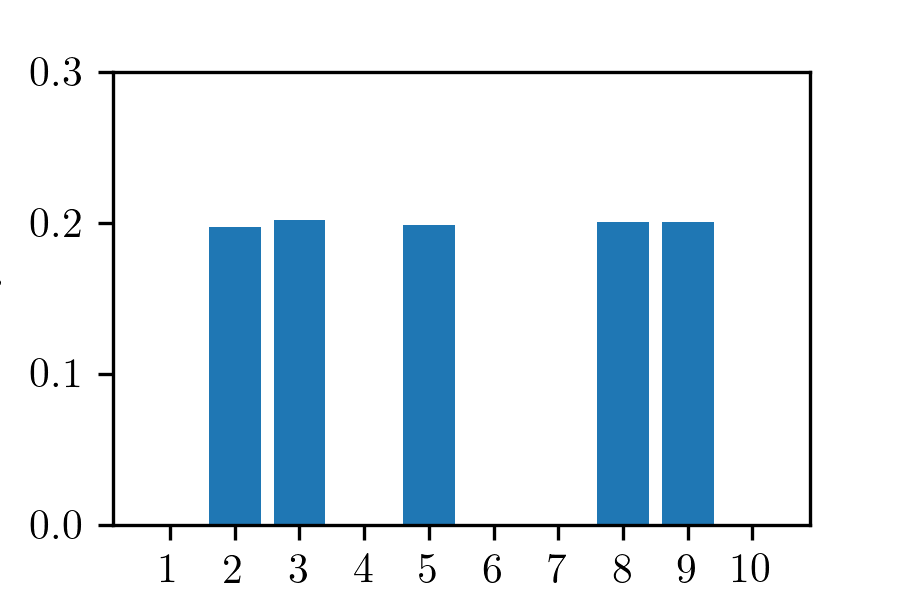}
 \end{subfigure}
    \caption{Estimated $\lambda_l$ for the $10$ latent parameterizations.}
     \end{subfigure}
  \caption{Simulation shows the LSP model recovers the latent parameterizations in multi-view clustering. \label{fig:multiview_sim}}
 \end{figure}

We first use a simulation to assess the multi-view clustering performance, under a large $V\gg n$. We use $V=50,000$ views, where each view has $n=150$. To produce distinct clustering patterns, we simulate $5$ different $W^{(l)}$'s, with each row generated from a Dirichlet distribution in a $3$-element simplex; for a better visualization, the rows are re-ordered and grouped by the index $\arg_k\max w^{(l)}_{i,k}$ (the most probable cluster). Then in each view, we randomly choose one of the five matrices (denoting the choice by $x_0^{(v)}\in\{1,\ldots,5\}$) as the parameterization, and sample the cluster labels $c^{(v)}_i\mid x^{(v)}=l \sim \text{Categorical} ( w^{(l)}_{i,1},w^{(l)}_{i,2},w^{(l)}_{i,3})$. Lastly, we generate each data point $y_i^{(v)} \mid c^{(v)}_i=k \sim N( \mu_k, I_2 )$, with $\mu_1=(0,0)'$, $\mu_2=(2,2)'$ and $\mu_3=(-2,-2)'$, so that there is moderate overlap between clusters. Figure~\ref{fig:multiview_sim}(a) plots five similarity matrices representative for those distinct patterns.

 When fitting the LSP model, we use $d=g=10$.  To examine the quality of initialization, we compare the initialized $\hat x^{(v)}$ and the oracle $x^{(v)}_0$: they show a high NMI at $0.83$, indicating that that the K-means on the log-odds of $S^{(v)}$'s gives a very good initialization. After running the EM algorithm, we do see $5$  of the $10$ estimated $\lambda_l$'s are shrunk to near zero [Figure~\ref{fig:multiview_sim}(c)], and the estimated $W^{(l)}W^{(l)\rm T}$'s with non-trivial $\lambda_l$'s indeed recover the five patterns [Figure~\ref{fig:multiview_sim}(b)]. For these high dimensional data, the algorithm takes about 10 minutes to finish on a CUDA GPU with 11Gb of memory.

\subsubsection{Clustering UCI Hand Written Digits}
Since the software is not readily available for most of the existing multi-view clustering methods, we use a dataset that was previously used for benchmark and reported by \cite{kumar2011co}. The dataset is the UCI Dutch utility maps handwritten digits data (\href{https://archive.ics.uci.edu/ml/datasets/Multiple+Features}{https://archive.ics.uci.edu/ml/datasets/Multiple+Features}), and does not require specific data processing; hence it can provide a fair comparison. The data have six views: (1) 76 Fourier coefficients of the character shapes, (2) 216 profile correlations, (3) 64 Karhunen-Loève coefficients, (4) 240-pixel averages in 2$\times$ 3 windows,  (5) Zernike moments and (6) 6 morphological features. For each digit, there are $200$ samples; the NMI is calculated as an evaluating criterion.

We compare our model with two other methods: using single-view spectral clustering independently in each view (SV-SC), and co-regularized spectral clustering (C-SC) \citep{kumar2011co}. When fitting the LSP model,we use $d=V$ as its possible max value, and $g=10$ as the known ground truth. The model converges to $2$ effective parameterizations: $\hat x^{(v)}=1$ for $v=1,\ldots,5$ and $\hat x^{(6)}=2$.

\begin{table}[H]
\centering
    \begin{tabular}{ l | c c c c c c}
            \hline
    Single View &   1 & 2 & 3 & 4 & 5 & 6  \\
        \hline
    SV-SC   &   0.571 & 0.618 & 0.646 & 0.635 & 0.523 & {\bf 0.474} \\
    LSP                  & {\bf 0.697} &   {\bf 0.706} & {\bf 0.697} & {\bf 0.705} & {\bf 0.705} & {\bf 0.474}\\
    \hline
    \end{tabular}
  \begin{tabular}{ l c c c}
            \hline
    Combining Multiple Views &    \\
        \hline
    SV-SC (feature concat)   &   0.619   \\
    C-SC  & {\bf 0.768} \\
    LSP consensus & 0.742\\
    \hline
    \end{tabular}
    \caption{Normalized mutual information for clustering UCI hand-written digits. Single view spectral clustering (SV-SC) and co-regularized spectral clustering (C-SC) are included for comparison. The NMIs for C-SC are obtained from \cite{kumar2011co}. \label{tb:UCI_digits}}
\end{table}

\begin{figure}[H]
 \begin{subfigure}[t]{0.45\textwidth}
 \centering
       \includegraphics[width=1\linewidth]{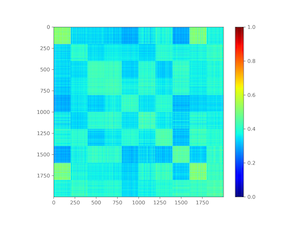}
             \caption{$P^{(v)}$ for the $1$st to the $5$th  views.}
 \end{subfigure}
\begin{subfigure}[t]{0.45\textwidth}
 \centering
       \includegraphics[width=1\linewidth]{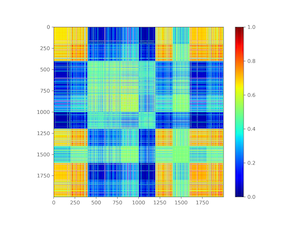}
             \caption{$P^{(v)}$ for the $6$th view.}
 \end{subfigure}
  \caption{Uncertainty for clustering hand written digit data, shown in the two estimated co-assignment probability matrices. The rows and columns are ordered according to the true labels of the digits. \label{fig:digit_uq}}
 \end{figure}

We first compute the point estimates $\hat c^{(v)}_i$ under each view. Table~\ref{tb:UCI_digits} shows that, compared to SV-SC, our model produces higher NMI in almost every view (except for the $6$th view). This is likely due to the sharing of information among the first few views in the LSP model. Then we combine views to produce a consensus. LSP has a better performance than using SV-SC on the concatenated features from all views, while it is slightly worse compared to C-SC. On the other hand, LSP has a unique advantage in the uncertainty quantification for each view. As shown in Figure~\ref{fig:digit_uq}, in the first parameterization, the main source of uncertainty is due to the overlap between the $1$st and the $9$th clusters; whereas the second parameterization has larger overlap among clusters $1,2,7,9$ and $10$.

\subsubsection{Clustering Brains via RNA-Sequencing Data}
We now consider a scientific application with the RNA Sequencing data originated from the human aging, dementia, and traumatic brain injury (TBI) study. The data are obtained from the Allen Institute for Brain Science \citep{miller2017neuropathological} (\href{https://aging.brain-map.org/download/index}{https://aging.brain-map.org/download/index}), and the hippocampus region is chosen for its important role in aging-related disease. Among the $107$ brains, there are $n=94$ containing gene expression data in the hippocampus. The age of the subjects at death has an average of $90$ and a standard deviation of $6$. There are $50,281$ genes, each with normalized gene-level FPKM values. Since most of the genes contain very little discriminability, a screening step is first carried out: the genes are ranked by their standard deviation divided by the median, with the top $V=1,000$ chosen for the downstream modeling.

This experiment treats each gene as a view, and clusters the $94$ brains using the gene expression. The LSP model was initialized at $d=30$ and $g=30$. It converges to $10$ effective latent parameterizations and at most $4$ non-trivial clusters.

\begin{figure}[H]
\begin{subfigure}[t]{.5\textwidth}
\centering
      \includegraphics[width=1\linewidth, height=1.8in]{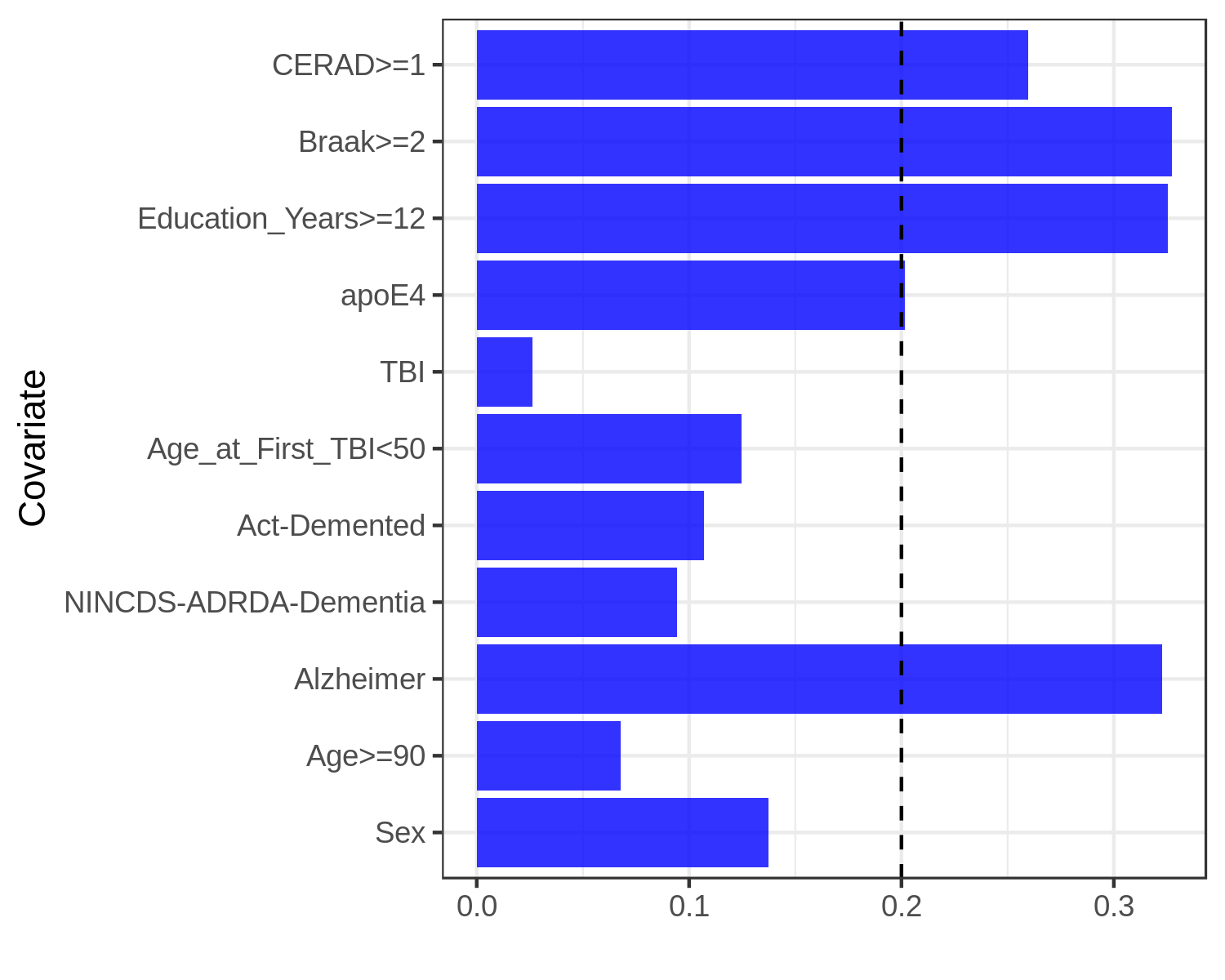}
            \caption{Maximum NMI over $1,000$ genes, comparing the clustering labels estimated from the LSP model and each clinical covariate.}
\end{subfigure}
\rulesep
\begin{subfigure}[t]{.5\textwidth}
\centering
      \includegraphics[width=1\linewidth, height=1.8in]{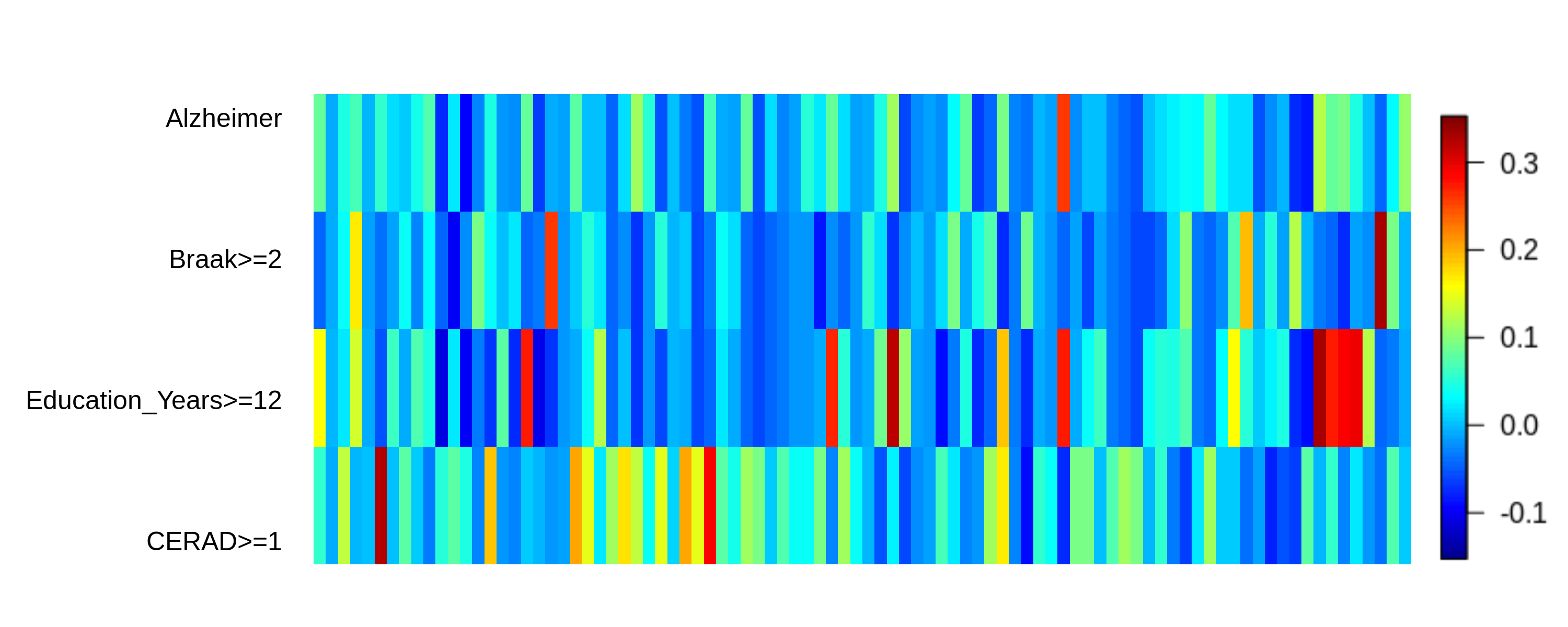}
            \caption{The NMIs of the $100$ selected genes associated with the four covariates.}
\end{subfigure}
 \caption{Clustering aging brains using gene expression. The results are compared with $11$ clinical covariates using NMIs. \label{fig:maxari}}
\end{figure}

For validation, the multi-view results are compared against $11$ observed clinical covariates --- such as sex, whether had TBI before, dementia evaluation scores, etc. For each clinical covariate, the NMIs are computed by comparing it against the estimated $\hat c^{(v)}_i$ for $v=1,\ldots,1000$. To see if there is a possible link between the gene(s) and clinical covariates, for each covariate, we take the maximum of $1000$ NMIs (MaxNMI) and plot it in Figure~\ref{fig:maxari}(a). As this involves multiple comparisons, to show the findings are unlikely to be false positives, we consider two additional baseline MaxNMIs: 1) we randomly draw $1000$ Bernoulli random vectors, each of length $n=94$, and compute the MaxNMI to each covariate. We repeat the simulation with different Bernoulli probabilities (ranging from $0.1$ to $0.9)$, the largest MaxNMI is $0.15$; 2)  it was previously reported in a meta-analysis study \citep{tan2016human}, that the difference in the covariate sex has no clear effect on the hippocampus area, in our experiment, it has MaxNMI $0.13$. Therefore, we choose MaxNMI$\ge 0.2$ as a cut-off.  For clarification, these results are mainly exploratory due to the small $n$; more data are needed before making any statistical claims.

Among all the covariates, the CERAD score (measuring the progression of Alzheimer's disease), the Braak stage (measuring the progression of Parkinson's disease and Alzheimer's disease) and the confirmed diagnose of Alzheimer's disease appear linked to a subset of gene expression. For the covariates seemingly less relevant to gene expression, the length of education also shows large MaxNMI, whereas the experience of traumatic brain injury (TBI), aging and dementia-related score show surprisingly low NMIs. Besides the MaxNMI, Figure~\ref{fig:maxari}(b) plots the NMIs of the top 100 genes associated with the selected four covariates. Clearly, each covariate seems more correlated with a distinct set of views/genes.



\section{Discussion}

In this article, we propose a method to directly estimate the cluster assignment probabilities for each data under multiple views. There are several interesting extensions that could be pursued. 1) The similarity matrix can be computationally prohibitive to handle when $n$ is large; therefore, a random feature map \citep{rahimi2008random} can be considered. A similar solution has recently been proposed for spectral clustering using random binning features \citep{wu2018scalable}. 2) We have assumed the views are given; in practice, if they are not known, we could use some domain-specific knowledge to estimate views. For example, in image processing, we could use edge detection and convolution of pixels to form each view. It is useful to study how they would impact the clustering results. 3) It is interesting to combine the LSP model with another loss function such as the one from a regression task, as an extension to our PAC-Bayes theory result. This could create generalized Bayes models and form new insights about the semi-supervised learning.

\appendix
\section{Proof of the main theorem}
\begin{proof}
Let $\text{KL}(P|| S)$ be the Kullbeck-Leibler divergence between two generating distributions for a cluster graph using $P$ or $S$ as the co-assignment probability matrix. The sample space of the distribution is a sub-space  $\mathcal{Z}\subset\{0,1\}^{n(n-1)/2}$ subject to constraints described in \eqref{eq:seq_prob_distribution}. We assume that $S$ is given.

Step 1. Change of measure:
\begin{equation*}
  \begin{aligned}
M    kl &\bigg[
\frac{1}{M}
\sum_{v=1}^M
\mathbb{E}_{\hat Z^{(v)}
\sim P}    loss \left(Z_0^{(v)}, \hat Z^{(v)}\right),
\mathbb{E}_{Z_0^{(v)}\sim \Pi_0}   \mathbb{E}_{\hat Z^{(v)}
\sim P}  loss \left(Z_0^{(v)}, \hat Z^{(v)}\right)\bigg] \\
&=M    kl \bigg[
\frac{1}{M}
\sum_{v=1}^M
\mathbb{E}_{\hat Z
\sim
P}    loss \left(Z_0^{(v)}, \hat Z\right),
\mathbb{E}_{Z_0^{(v)}\sim \Pi_0}   \mathbb{E}_{\hat Z
\sim P}  loss \left(Z_0^{(v)}, \hat Z^{}\right)\bigg] \\
    & \le   \mathbb{E}_{\hat Z
\sim P}    M kl \bigg[
\frac{1}{M}
\sum_{v=1}^M
loss \left(Z_0^{(v)}, \hat Z\right),
\mathbb{E}_{Z_0^{(v)}\sim \Pi_0}   loss \left(Z_0^{(v)}, \hat Z^{}\right)\bigg] \\
    & \le   \text{KL} (P||S ) + \log \mathbb{E}_{\hat Z \sim S} \exp\bigg \{ M kl \bigg[
\frac{1}{M}
\sum_{v=1}^M
loss \left(Z_0^{(v)}, \hat Z\right),
\mathbb{E}_{Z_0^{(v)}\sim \Pi_0}   loss \left(Z_0^{(v)}, \hat
Z^{}\right)\bigg] \bigg\},
  \end{aligned}
\end{equation*}
where the first equality is because $\hat Z^{(v)}$'s are iid, the first inequality uses the convexity of the $kl$ function; the second inequality uses the change of measure inequality [Lemma 4 in \citep{seldin2010pac}].

Due to the Markov's inequality, with probability greater than $1-\delta$,
\begin{equation*}
        \begin{aligned}
    &\mathbb{E}_{\hat Z \sim S} \exp\bigg \{
M kl \bigg[
\frac{1}{M}
\sum_{v=1}^M
loss \left(Z_0^{(v)}, \hat Z\right),
\mathbb{E}_{Z_0^{(v)}\sim \Pi_0}   loss \left(Z_0^{(v)}, \hat
Z^{}\right)\bigg] \bigg\} \\
     & \le       \frac{1}{\delta}  \mathbb{E}_{Z_0^{(v)}\stackrel{iid}{\sim} \Pi_0}     \mathbb{E}_{\hat Z \sim S} \exp\bigg \{
M kl \bigg[
\frac{1}{M}
\sum_{v=1}^M
loss \left(Z_0^{(v)}, \hat Z\right),
loss \left(Z_0^{(v)}, \hat
Z^{}\right)\bigg] \bigg\} \\
     & =       \frac{1}{\delta}   \mathbb{E}_{\hat Z \sim S}
\mathbb{E}_{Z_0^{(v)}\stackrel{iid}{\sim} \Pi_0}    \exp\bigg
\{
M kl \bigg[
\frac{1}{M}
\sum_{v=1}^M
loss \left(Z_0^{(v)}, \hat Z\right),
loss \left(Z_0^{(v)}, \hat
Z^{}\right)\bigg] \bigg\} ,
        \end{aligned}
\end{equation*}
and the last equality is due to $kl(.)$ is upper bounded, hence we can use Fubini's theorem.

Step 2. Bounding the exponential $kl$ function by a constant:

 Using Theorem 1 of \citep{maurer2004note},
\begin{equation*}
        \begin{aligned}
     &\mathbb{E}_{Z_0^{(v)}\stackrel{iid}{\sim} \Pi_0}    \exp\bigg
\{
M kl \bigg[
\frac{1}{M}
\sum_{v=1}^M
loss \left(Z_0^{(v)}, \hat Z\right),
loss \left(Z_0^{(v)}, \hat
Z^{}\right)\bigg] \bigg\}   \\
     &\le \exp(\frac{1}{12M}) \sqrt{\frac{\pi M}{2}} + 2.
        \end{aligned}
\end{equation*}

Step 3. Relaxing the $\text{KL}$-divergences between two cluster graph distributions to total element-wise $kl$ divergences:

\begin{equation*}
        \begin{aligned}
     \text{KL} (
P||S) &=  \underset{Z\in \mathcal Z} \sum   \{ 1(z_{i,j}=1) { p_{i,j}}  \log \frac{ p_{i,j}}{s_{i,j}}   \\
     &+ 1(z_{i,j}=0)  (1-{p_{i,j}})\log \frac{1- p_{i,j}}{1-s_{i,j}} \} \\
&  \le    \underset{Z\in \{0,1\}^{n(n-1)/2}} \sum   \{ 1(z_{i,j}=1) {  p_{i,j}}  \log \frac{p_{i,j}}{s_{i,j}}   \\
     &+ 1(z_{i,j}=0)  (1-{p_{i,j}})\log \frac{1- p_{i,j}}{1-s_{i,j}} \}\\
& = \sum_{j<i}  kl(  p_{i,j} || s_{i,j}),
        \end{aligned}
\end{equation*}
where the inequality is due to the non-negativity of each $kl$ function and $\mathcal{Z} \subseteq \{0,1\}^{n(n-1)/2}$.

Combining the results,
$$
    KL \big[ R(\Pi_0, \phi) \;||\; R(\Pi_M, \phi)\big]
    \le  \frac{
  \sum_{j<i}  KL( p_{i,j}|| s_{i,j}) + \log \{\exp(\frac{1}{12M})
\sqrt{\frac{\pi M}{2}} + 2\} - \log \delta
      }{M}.
$$
Taking $S=S^{(v)}$, summing both sides over $v=1,\ldots,M$ and dividing by $M$ yields the result.
\end{proof}

\section{Single View Simulation}
\begin{figure}[H]
 \begin{subfigure}[t]{0.32\textwidth}
 \centering
       \includegraphics[width=1\linewidth]{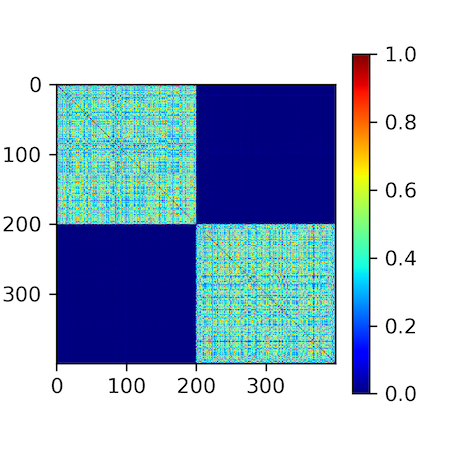}
             \caption{$S$ for (a).}
 \end{subfigure}
\begin{subfigure}[t]{0.32\textwidth}
 \centering
       \includegraphics[width=1\linewidth]{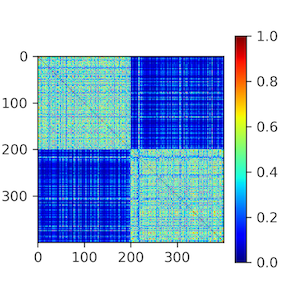}
             \caption{$S$ for (b).}
 \end{subfigure}
 \begin{subfigure}[t]{0.32\textwidth}
 \centering
       \includegraphics[width=1\linewidth]{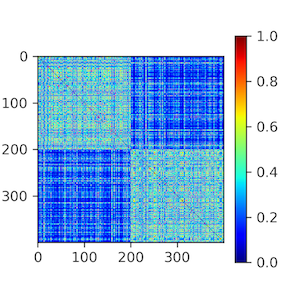}
             \caption{ $S$ for (c).}
 \end{subfigure}
  \begin{subfigure}[t]{0.32\textwidth}
 \centering
       \includegraphics[width=1\linewidth]{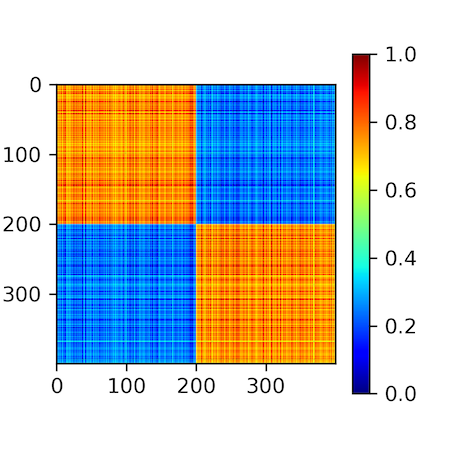}
             \caption{Estimated $P$ for (a).}
 \end{subfigure}
\begin{subfigure}[t]{0.32\textwidth}
 \centering
       \includegraphics[width=1\linewidth]{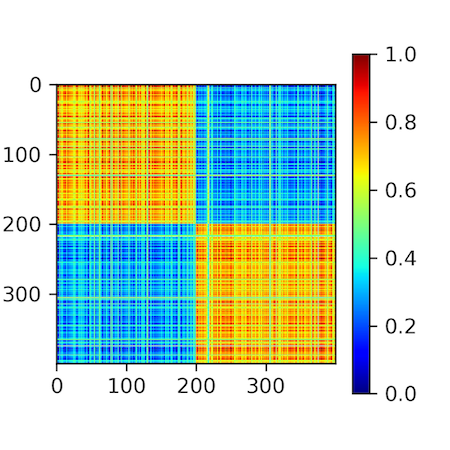}
       \caption{Estimated $P$ for (b).}
      \end{subfigure}
 \begin{subfigure}[t]{0.32\textwidth}
 \centering
       \includegraphics[width=1\linewidth]{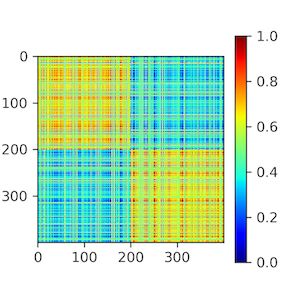}
              \caption{Estimated $P$ for (c).}
            \end{subfigure}\\
  \begin{subfigure}[t]{0.32\textwidth}
 \centering
       \includegraphics[width=1\linewidth]{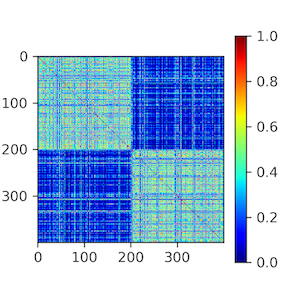}
       \caption{ $S$ for (d).}
 \end{subfigure}
 \begin{subfigure}[t]{0.32\textwidth}
 \centering
       \includegraphics[width=1\linewidth]{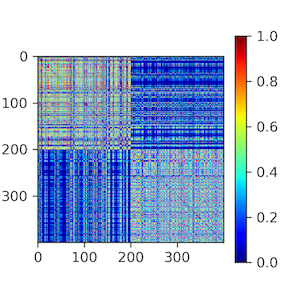}
       \caption{ $S$ for (e).}
 \end{subfigure}
\begin{subfigure}[t]{0.32\textwidth}
 \centering
       \includegraphics[width=1\linewidth]{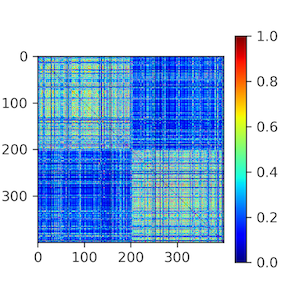}
       \caption{ $S$ for (f).}
 \end{subfigure}
  \begin{subfigure}[t]{0.32\textwidth}
 \centering
       \includegraphics[width=1\linewidth]{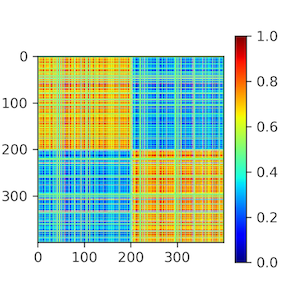}
       \caption{Estimated $P$ for (d).}
 \end{subfigure}
 \begin{subfigure}[t]{0.32\textwidth}
 \centering
       \includegraphics[width=1\linewidth]{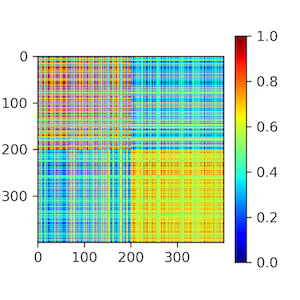}
       \caption{Estimated $P$ for (e).}
 \end{subfigure}
\begin{subfigure}[t]{0.32\textwidth}
 \centering
       \includegraphics[width=1\linewidth]{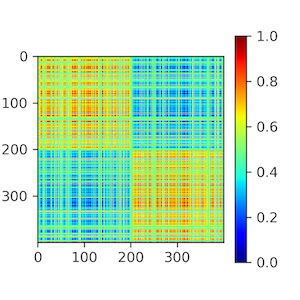}
       \caption{Estimated $P$ for (f).}
 \end{subfigure}
  \caption{Computed similarity matrices and estimated co-assignment probability matrices in the single view clustering experiments (Section 5.1).}
 \end{figure}

\bibliography{../reference}
\bibliographystyle{chicago}

\end{document}